\documentclass[journal]{IEEEtran} 
\usepackage{amsmath,amsfonts}
\usepackage{algorithmic}
\usepackage{algorithm}
\usepackage{array}
\usepackage[caption=false,font=normalsize,labelfont=sf,textfont=sf]{subfig}
\usepackage{textcomp}
\usepackage{stfloats}
\usepackage{url}
\usepackage{verbatim}
\usepackage{graphicx}
\usepackage{cite}
\usepackage{multirow}
\usepackage{xcolor}
\usepackage{soul}
\sethlcolor{yellow} 
\usepackage{siunitx}

\usepackage{scalerel}
\usepackage{tikz}
\usetikzlibrary{svg.path}

\definecolor{orcidlogocol}{HTML}{A6CE39}
\tikzset{
  orcidlogo/.pic={
    \fill[orcidlogocol] svg{M256,128c0,70.7-57.3,128-128,128C57.3,256,0,198.7,0,128C0,57.3,57.3,0,128,0C198.7,0,256,57.3,256,128z};
    \fill[white] svg{M86.3,186.2H70.9V79.1h15.4v48.4V186.2z}
                 svg{M108.9,79.1h41.6c39.6,0,57,28.3,57,53.6c0,27.5-21.5,53.6-56.8,53.6h-41.8V79.1z M124.3,172.4h24.5c34.9,0,42.9-26.5,42.9-39.7c0-21.5-13.7-39.7-43.7-39.7h-23.7V172.4z}
                 svg{M88.7,56.8c0,5.5-4.5,10.1-10.1,10.1c-5.6,0-10.1-4.6-10.1-10.1c0-5.6,4.5-10.1,10.1-10.1C84.2,46.7,88.7,51.3,88.7,56.8z};
  }
}

\newcommand\orcidicon[1]{\href{https://orcid.org/#1}{\mbox{\scalerel*{
\begin{tikzpicture}[yscale=-1,transform shape]
\pic{orcidlogo};
\end{tikzpicture}
}{|}}}}

\usepackage{amssymb}  
\usepackage{pifont}
\newcommand{\cmark}{\ding{51}}%
\newcommand{\xmark}{\ding{55}}%
\usepackage{comment}
\usepackage{tabularx}
\usepackage{tablefootnote}
\usetikzlibrary{tikzmark}
\usepackage{soul}

\usepackage{hyperref}

\begin{document}

\title{Challenges for Monocular\\6D Object Pose Estimation in Robotics}

\author{
Stefan Thalhammer\orcidicon{0000-0002-0008-430X}\,, \IEEEmembership{Member, IEEE}, 
Dominik Bauer\orcidicon{0000-0002-1260-1319}\,, \IEEEmembership{Member, IEEE}, 
Peter H{\"o}nig\orcidicon{0000-0002-8990-6110}\,,
\IEEEmembership{Student Member, IEEE},
Jean-Baptiste Weibel\orcidicon{0000-0003-0201-4740}\,, \IEEEmembership{Member, IEEE},
José García-Rodríguez\orcidicon{0000-0002-7798-3055}\,, \IEEEmembership{Senior Member,
IEEE}
and Markus Vincze\orcidicon{0000-0002-2799-491X}\,, \IEEEmembership{Senior Member, IEEE}
\thanks{Stefan Thalhammer is with the Industrial Engineering Departent, UAS Technikum Vienna, Austria
        {\tt\footnotesize thalhammer@technikum-wien.at}}%
\thanks{Peter H{\"o}nig, Jean-Baptiste Weibel, and Markus Vincze are with the Automation and Control Institute, TU Wien, Austria
        {\tt\footnotesize \{thalhammer,hoenig,weibel,vincze\}@acin.tuwien.ac.at}}%
\thanks{Dominik Bauer is with the Columbia Artificial Intelligence and Robotics Lab, Columbia University, USA
        {\tt\footnotesize dominik.bauer@columbia.edu}}%
\thanks{José García-Rodríguez is with the Department of Computer Technology, University of Alicante, Spain
        {\tt\footnotesize jgarcia@dtic.ua.es}}%
}




\maketitle

\begin{abstract}
Object pose estimation is a core perception task that enables, for example, object manipulation and scene understanding.
The widely available, inexpensive and high-resolution RGB sensors and CNNs that allow for fast inference make monocular approaches especially well suited for robotics applications.
We observe that previous surveys on establish the state of the art for varying modalities, single- and multi-view settings, and datasets and metrics that consider a multitude of applications.
We argue, however, that those works' broad scope hinders the identification of open challenges that are specific to monocular approaches and the derivation of promising future challenges for their application in robotics.
By providing a unified view on recent publications from both robotics and computer vision, we find that occlusion handling, pose representations, and formalizing and improving category-level pose estimation are still fundamental challenges that are highly relevant for robotics.
Moreover, to further improve robotic performance, large object sets, novel objects, refractive materials, and uncertainty estimates are central, largely unsolved open challenges.
In order to address them, ontological reasoning, deformability handling, scene-level reasoning, realistic datasets, and the ecological footprint of algorithms need to be improved.
\end{abstract}


\section{Introduction}
\begin{figure}[t!]
   \centering
    \includegraphics[width=0.75\columnwidth]{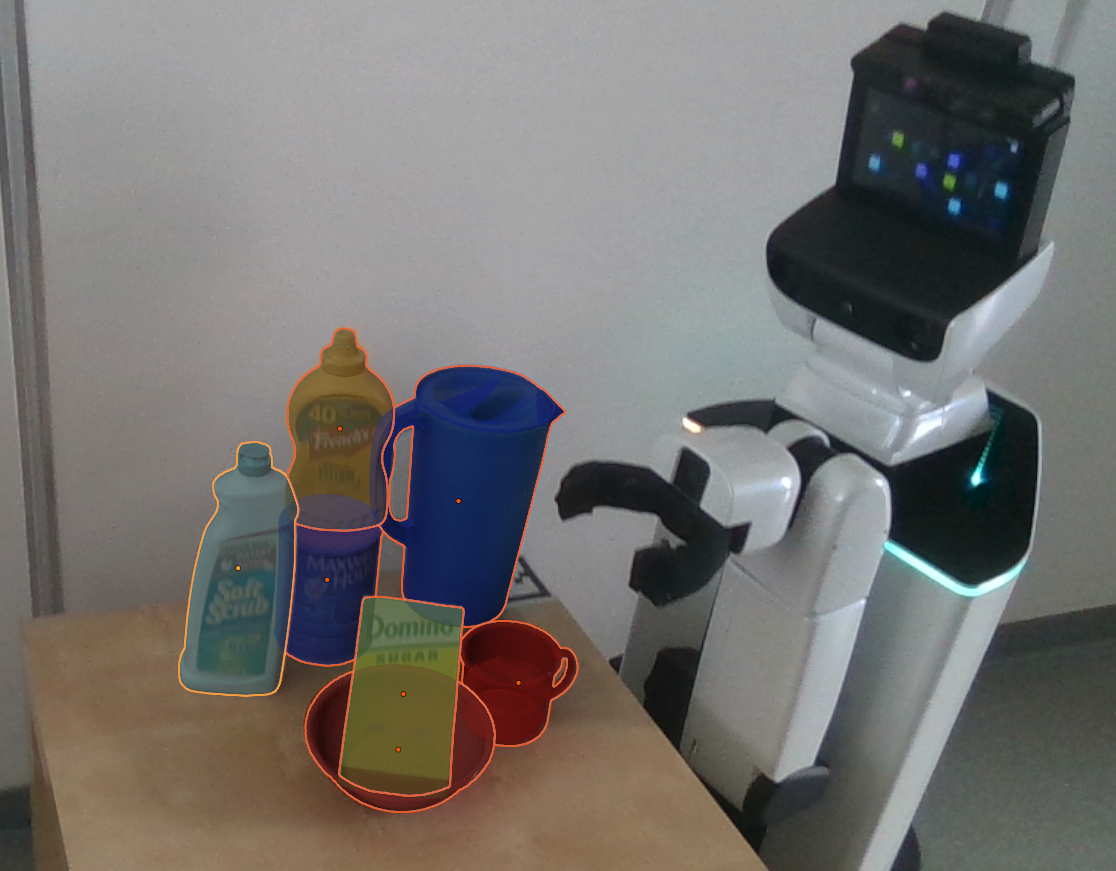}
   \caption{\textbf{Object poses} YCB-V~\cite{xiang2017posecnn} objects overlaid with their annotated poses.}
   \label{fig:annotated_poses}
   \vspace{-2ex}
\end{figure}

\IEEEPARstart{O}{bject} pose estimation is an essential task for applications like robotic grasping~\cite{patten2020dgcm}, bin picking~\cite{kleeberger2019large} and semantic scene understanding~\cite{Nie_2020_CVPR}.
Though depth readings have proven to be reliable for deriving poses of objects with distinctive geometry, special hardware is required.
These sensors are comparably expensive, have varying noise patterns at different distances, image regions and between sensors and have lower resolutions than RGB sensors.
RGB sensors have a wider range of applications due to their low cost, expressiveness, low noise compared to depth sensors, and efficient processing with CNNs.
They offer abundant information to disambiguate object symmetries through texture, provides noise-free captures of objects with challenging material properties like transparency, which are difficult to detect with depth sensors, and have larger working ranges ideally suited for the unstructured open-world.
As a consequence RGB gained significant relevance for object pose estimation in the recent years.
In the most recent edition of the Benchmark for $6D$ Object Pose estimation (BOP), the top two, and three out of the top five methods generate initial pose hypotheses in RGB, and only use depth data for pose refinement~\cite{hodan2022bop}.
This trend, that monocular approaches, taking RGB as input, outperform depth-based ones already started in the previous BOP edition~\cite{hodan2020epos}. 
Consequently, RGB approaches are the preferred choice for detecting objects and providing initial pose hypotheses.
This substantiates the importance of monocular single-shot $6D$ object pose estimation for robotics.
Figure~\ref{fig:annotated_poses} shows exemplary pose annotations.

Recent summaries of the state-of-the-art for object pose estimation~\cite{he20206d,sahin2020review,hoque2021comprehensive,fan2022deep,marullo20236d} report domain shift, end-to-end training, occlusion handling, symmetry handling, improving datasets, reasoning about geometric properties, clutter handling, and pose estimation of metallic, deformable, and textureless objects as future challenges.
Since the field is evolving rapidly, works published in the meantime largely solved the problems identified in~\cite{he20206d}, namely domain shift, handling symmetries and clutter, textureless objects, and end-to-end training on standard datasets.
At the same time, related surveys consider methods' performance on standard datasets, or their categorization into established taxonomies.
In~\cite{sahin2020review}, the authors present a categorization of pose estimators' accuracy using standard datasets and metrics, and in~\cite{hoque2021comprehensive}, a discussion of $3D$ object detection and $6D$ object pose estimation methods for autonomous vehicles is presented.  
The recent work of Fan et al.~\cite{fan2022deep} presents the state of the art for instance- and category-specific pose estimation and tracking methods on standard datasets.
Also recently, Marullo et al.~\cite{marullo20236d}, consider approaches along three main categories; template-based, feature-based methods, and learning-based methods.
In comparison to all of these works, we specifically focus on discussing \textit{monocular} object pose estimation for \textit{robotics}, with a pronounced emphasis on future developments.

We discuss its relevance for different robotic applications, and anticipate changes in the field due to recent advancements in generalization of robot learning~\cite{firoozi2023foundation,padalkar2023open,fang2023anygrasp,black2023zero}.
Additionally, we correlate developments in pose estimation with those in robotic manipulation and policy learning, deriving contact points between these fields that will influence future research.
Thereby, this review derives ongoing and future robotics-related perception challenges that received little attention in prior reviews.

We base our discussion on a review of current key problems, identified by extracting works published in the years $2021$, $2022$, $2023$, and $2024$ (up to paper submission) in robotics and computer vision venues with high scientific impact.
We consider works before that to be adequately summarized in previous surveys~\cite{he20206d} and the selected works from recent years to be more indicative of future challenges.
Our review of these works suggests that overcoming the domain shift under the closed-world assumption in industrial and household environments appears to be largely solved.
We argue that occlusion handling is a characteristic of approaches that generally provide accurate poses and handle outlier well.
We highlight the importance of more fundamental problems such as finding an unambiguous yet compact pose representation, multi-object handling, category-level and novel object pose estimation beyond known categories, handling challenging material properties, and the resulting visual uncertainties.
Future challenges are derived by identifying the gaps in this problem landscape from a robotics perspective. 
Downstream robotic and vision tasks that leverage knowledge of the object interaction on a scene-level, object ontologies that describe non-distinct class memberships, and language-grounded foundation models help resolve edge cases in the real world. 
Novel datasets that reflect the complexity of unstructured, natural robotic environments will highlight the improvement required beyond current standard algorithms, and progress robotic research. 
Measures need to be taken to reduce the environmental impact of pose estimation research and its application in robotics.
Finally, recent advancements in generalist robot policies and grasp generalization, and foundation models, are bound to cause a paradigm shift for object manipulation and scene understanding.
In summary, this review provides
\begin{itemize}
    \item arguments which indicate that many core challenges identified by previous surveys, such as domain adaptation, are thoroughly investigated in the meantime,
    \item an evaluation of the relevance of common dataset for open challenges in robotics, and
    \item a survey focused on robotics-specific challenges in ever-changing unstructured environments, allowing us to a) identify solutions to, e.g., handling many and novel objects, that still lack in performance for robotic application and b) identify problems that occur in typical robotics settings, such as non-distinct categories or natural scene structures, that are overlooked by previous surveys.
\end{itemize}

The remainder of the paper is organized as follows.
Section~\ref{sec:problems}, presents our analyses of currently investigated research problems. Sections~\ref{sec:datasets} and~\ref{sec:trends} presents common datasets and the status of the currently progress toward solving these research problems. 
Subsequently, Section~\ref{sec:future} presents promising novel directions. 
Section~\ref{sec:conclusion} provides a summary of our findings and discusses future work.

\section{Ongoing Research and Common Datasets}
\label{sec:problems}

In order to investigate problems, a representative sample of works published in the years $2021$, $2022$, $2023$ and $2024$, up to the first draft submission of this script, was gathered.
Publications are selected from the top six robotics and computer vision venues, as determined by the h5-index, impact factor and SJR metric. 
For robotics we consider works published in IEEE/RAS ICRA, IEEE T-RO and RA-L, IEEE/RSJ IROS, IEEE/ASME T-MECH, and SAGE IJRR.
For computer vision we consider works published in IEEE/CVF CVPR and ICCV, Springer ECCV, IEEE T-PAMI and T-IP, and Elsevier PR.

Table~\ref{tab:topics} presents the motivating problems for the respective scientific works, sorted by frequency. The total number of publications considered for identifying these active research problems is $72$, of which $33$ are published in the field of robotics and $39$ in computer vision. Single works count toward multiple problems. 
The following section presents datasets that are commonly used for benchmarking these single-shot approaches for $6D$ object pose estimation.

\begin{table}
\caption{\textbf{Ongoing monocular object pose estimation research} Frequencies of the motivating problems for papers published in the years 2021, 2022, 2023, and 2024 (up to paper submission).}
\centering
\begin{tabular}{|c|c|c|c|}
 \hline
 Problem & References & freq. \\ \hline
 Domain shift & \cite{zhang2021keypoint}\cite{hu2022perspective} \cite{nguyen2022templates} \cite{wang2021self6d++} \cite{thalhammer2021pyrapose} \cite{shi2021fastUQ} \cite{lu2022slam} & \\
 & \cite{ikeda2022sim2real} \cite{yang2022image} \cite{jawaid2023towards} \cite{Chen_2023_CVPR} & 11 \\
 Occlusion & \cite{ma2022robust} \cite{nguyen2022templates} \cite{peng2022pvnet} \cite{shugurov2022dpodv2} \cite{liu2021mfpn6d} \cite{liu2021kdfnet} \cite{mei2022spatial} & \\
 &\cite{hai2023rigidity} \cite{xu20236d} \cite{wang2022multiple} & 10 \\
 Representation & \cite{huang2022neural} \cite{di2021so} \cite{park2022dprost} \cite{su2022zebrapose} \cite{yen2021inerf} \cite{liu2023linear} \cite{lian2023checkerpose} & \\
 & \cite{li2024mrc} \cite{yang2023exploring} & 9 \\
 Novel objects & \cite{liu2022gen6d} \cite{shugurov2022osop} \cite{nguyen2022templates} \cite{gu2022ossid} \cite{saxena2023generalizable} \cite{moon2024genflow} \cite{nguyen2023nope} & \\
 & \cite{wen2024foundationpose} \cite{ausserlechner2023zs6d} & 9 \\
 End-to-end & \cite{wang2021gdr} \cite{di2021so} \cite{chen2022epro} \cite{lipson2022coupled} \cite{cao2022dgecn} \cite{hai2023shape} \cite{liu2023linear} & 7 \\
 Robotic application & \cite{hu2021wide} \cite{viviers2024advancing} \cite{sapienza2023underwater} \cite{tang2024rov6d} \cite{sun2023panelpose} \cite{ulmer2024orbital} \cite{monguzzi2024cable} & 7 \\
 Refinement & \cite{iwase2021repose} \cite{haugaard2021surfemb} \cite{lipson2022coupled} \cite{shugurov2022dpodv2} \cite{araki2021iterative} \cite{moon2024genflow} & 6 \\
 Symmetry & \cite{wen2022disp6d} \cite{haugaard2021surfemb} \cite{richter2021handling} \cite{bengston2021pose} \cite{jeon2023ambiguity} \cite{li2024mrc} & 6 \\
 Multi-object & \cite{iwase2021repose} \cite{wen2022disp6d} \cite{thalhammer2021pyrapose} \cite{jeon2023ambiguity} \cite{viviers2024advancing} \cite{chaitanya2022physics} & 6 \\
 Category & \cite{fan2022object} \cite{wen2022disp6d} \cite{ma2022robust} \cite{lin2022single}  \cite{Lee2021category}  & 5 \\ 
 Material properties & \cite{he2022generative} \cite{liu2021mfpn6d} \cite{yu2023TGFnet} \cite{he2023contour} \cite{he2024ggop} & 5 \\
 Uncertainty estimation & \cite{Yang_2023_CVPR} \cite{jeon2023ambiguity} \cite{shi2021fastUQ} & 3 \\
 Learning principles & \cite{guo2023knowledge} \cite{Chen_2023_CVPR} \cite{hai2023pseudo} & 3 \\
\hline
\end{tabular}
\label{tab:topics}
   \vspace{-1ex}
\end{table}

\subsection{Datasets}
\label{sec:datasets}

Common datasets are crucial for comparing different methods in a reproducible manner.
As a consequence pose estimation methods are designed with the focus of maximizing accuracy on them. 
Obviously, if the datasets' complexity does not reflect the relevant real-world complexity of robotic scenarios, this will lead to a performance discrepancy at deployment.
The following discussion of these datasets thus helps identifying shortcomings in the problem landscape of mainstream methods, and allows deriving important challenges that are required to improve dataset design to the performance in the real world. 
Datasets that do not provide real test images are not included in the following list, since methods tested on them do not reflect the expected performance in the real world and thus in robotics applications.

We first present instance-level datasets, which are the most used and considered the de-facto standard for benchmarking object pose estimation approaches~\cite{hinterstoisser2012model,brachmann2014learning,hodan2018bop,doumanoglou2016recovering,homebrewedDB,hodan2017tless,drost2017introducing}.
Following that, category-level datasets are listed. These are of interest for robotics due to the possibility of testing the generalization ability of approaches~\cite{wang2019normalized,chen2022clearpose,dai2022dreds,wang2022phocal,liu2020keypose}.
Lastly, we discuss datasets that include annotations for grasping, used to benchmark the object pose estimation works presented in~\cite{jung2022housecat6d,fang2022transcg,Cao2021suction}.

\begin{figure*}[t!]
   \centering
    \includegraphics[width=1.9\columnwidth]{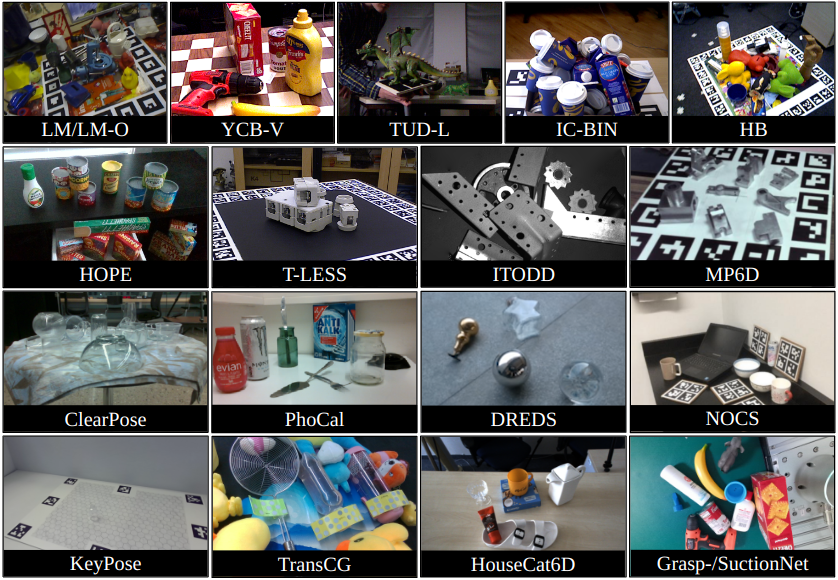}
   \caption{\textbf{Datasets overview} Exemplary images of well established datasets and novel ones that provide diverse important challenges.}
   \label{fig:instdata}
      \vspace{-2ex}
\end{figure*}

\begin{table*}
\caption{\textbf{Dataset comparison} Challenges present in each datasets. Note that no single dataset provides full coverage.} 
\centering
\begin{tabular}{|l|c|c|c|c|c|c|c|}
\hline
 Dataset & Multi-object & Multi-instance & Occlusion & Category-level & Symmetries & Challenging Materials \\ \hline
1) LM/LM-O\cite{hinterstoisser2012model,brachmann2014learning} & \color{green}\cmark & \color{red}\xmark & \color{green}\cmark & \color{red}\xmark & \color{green}\cmark & \color{red}\xmark \\
 2) YCB-V\cite{xiang2017posecnn} & \color{green}\cmark & \color{red}\xmark & \color{green}\cmark & \color{red}\xmark & \color{green}\cmark & \color{red}\xmark \\
 3) TUD-L\cite{hodan2018bop} & \color{red}\xmark & \color{red}\xmark & \color{red}\xmark & \color{red}\xmark & \color{red}\xmark & \color{red}\xmark \\
 4) IC-BIN\cite{doumanoglou2016recovering} & \color{green}\cmark & \color{green}\cmark & \color{green}\cmark & \color{red}\xmark & \color{green}\cmark & \color{red}\xmark \\
 5) HB\cite{homebrewedDB} & \color{green}\cmark & \color{red}\xmark & \color{green}\cmark & \color{red}\xmark & \color{green}\cmark & \color{red}\xmark \\
 6) HOPE\cite{tyree20226} & \color{green}\cmark & \color{green}\cmark & \color{green}\cmark & \color{red}\xmark & \color{red}\xmark & \color{red}\xmark \\
 7) T-LESS\cite{hodan2017tless} & \color{green}\cmark & \color{green}\cmark & \color{green}\cmark & \color{red}\xmark & \color{green}\cmark & \color{red}\xmark \\
 8) ITODD\cite{drost2017introducing} & \color{green}\cmark & \color{green}\cmark & \color{green}\cmark & \color{red}\xmark & \color{green}\cmark & \color{green}\cmark \\
 9) MP6D\cite{chen2022MP6D} & \color{green}\cmark & \color{red}\xmark & \color{green}\cmark & \color{red}\xmark & \color{green}\cmark & \color{green}\cmark \\
 10) ClearPose\cite{chen2022clearpose} & \color{green}\cmark & \color{red}\xmark & \color{green}\cmark & \color{red}\xmark & \color{green}\cmark & \color{green}\cmark \\
 11) NOCS\cite{wang2019normalized} & \color{green}\cmark & \color{green}\cmark & \color{green}\cmark & \color{green}\cmark & \color{green}\cmark & \color{red}\xmark \\
 12) PhoCal\cite{wang2022phocal} & \color{green}\cmark & \color{red}\xmark & \color{green}\cmark & \color{green}\cmark & \color{green}\cmark & \color{green}\cmark \\
 13) DREDS\cite{dai2022dreds} & \color{green}\cmark & \color{red}\xmark & \color{green}\cmark & \color{green}\cmark & \color{green}\cmark & \color{green}\cmark \\
 14) KeyPose\cite{liu2020keypose} & \color{red}\xmark & \color{red}\xmark & \color{red}\xmark & \color{green}\cmark & \color{green}\cmark & \color{green}\cmark \\
 15) TransCG\cite{fang2022transcg} & \color{green}\cmark & \color{red}\xmark & \color{green}\cmark & \color{green}\cmark & \color{green}\cmark & \color{green}\cmark \\
 16) HouseCat6D\cite{jung2022housecat6d} & \color{green}\cmark & \color{red}\xmark & \color{green}\cmark & \color{green}\cmark & \color{green}\cmark & \color{green}\cmark \\
 17) Grasp/-SuctionNet\cite{fang2020graspnet,Cao2021suction} & \color{green}\cmark & \color{red}\xmark & \color{green}\cmark & \color{red}\xmark & \color{green}\cmark & \color{red}\xmark \\ \hline
\end{tabular}
\label{tab:datasets}
   \vspace{-1ex}
\end{table*}

\subsubsection{Linemod/-Occlusion~\cite{hinterstoisser2012model,brachmann2014learning} (LM/LM-O)} 
$13$ objects with one test sequence of approximately $1,300$ images for each are provided. 
LM has been introduced as challenging due to clutter backgrounds, yet objects are fully visible in virtually all test images. 
To create challenging occlusion patterns,~\cite{brachmann2014learning} re-annotated LM's second test sequence, introducing Linemod-Occlusion (LM-O).
LM-O provides annotations for eight of LM's objects under severe occlusion.
For both, no real-world training images are available, yet an established standard is that $15\%$ of the test images of LM are used for training~\cite{Park_2019_ICCV}. 

\subsubsection{YCB-video (YCB-V)~\cite{xiang2017posecnn}} $21$ objects from YCB~\cite{calli2015ycb} are annotated in $134k$ real images.
Out of the complete set, $113k$ images are used for training and the rest for testing.
Challenges are varying levels of occlusion and large variety in illumination.

\subsubsection{TUD-L~\cite{hodan2018bop}} This dataset was presented alongside the $2018$ edition of the BOP.
In the current edition of the challenge, three objects with $11,000$ real-world training and $200$ test images for each, are used.
The challenges are strong viewpoint and illumination changes.
Additionally, in contrast to other datasets, the objects are not supported by a table plane, only a small object-sized support.

\subsubsection{IC-BIN~\cite{doumanoglou2016recovering}} This dataset provides multi-object test images of eight objects in clutter and with occlusion.
For the BOP, three sequences featuring three of the eight objects are used, resulting in $150$ test images.
The test set features tightly packed objects in a pile, resulting in heavy occlusion.

\subsubsection{Homebrewed-Database (HB)~\cite{homebrewedDB}} This dataset consists of $33$ household and industrial objects.
No official training set is available. Of each of the available $13$ sequences, $340$ images are used for validation.
Three sets with $100$ images each are used as the BOP test set.
The challenges of the dataset are symmetries, occlusion and strong illumination variation.

\subsubsection{HOPE~\cite{tyree20226}} The dataset consists of $28$ toy grocery objects in household scenes.  
For testing, $238$ images in $50$ different scenes are provided.
The challenges are multiple object instances per image, occlusion, strong illumination changes, and clutter. 

\subsubsection{T-LESS~\cite{hodan2017tless}} Provided are $30$ highly symmetrical texture-less objects, similar in shape, and some objects are parts of others; objects one might find in an industrial setting.
For each of the dataset objects, CAD models, reconstructions, and $1,296$ real training images of uniformly distributed views are available.
The challenges are handling symmetries, occlusion, textureless objects, and inter-object similarities.

\subsubsection{ITODD~\cite{drost2017introducing}} consists of $28$ metallic industrial objects and grayscale images. 
This dataset is also part of the BOP, yet it is the only dataset in the benchmark with more challenging material properties than diffusely reflective surfaces.
The test set exhibits heavy occlusion and challenging material properties for both RGB and depth-based methods.

\subsubsection{MP6D~\cite{chen2022MP6D}} provides $20,100$ real images, where $20$ object instances are annotated.
Of these, $6k$ are used for testing.
The fully specular-reflective industrial objects are placed in a multi-object setting.
Aside from their reflectivity, the challenges are handling clutter, symmetry, and occlusion.

\subsubsection{ClearPose~\cite{chen2022clearpose}} The dataset features $63$ transparent objects, most of them exhibiting symmetries.
Provided are $354,481$ real images in $51$ scenes with diverse backgrounds, changing illumination and clutter with occlusion.

\subsubsection{NOCS~\cite{wang2019normalized}} presents two datasets for category-level object pose estimation.
Context-Aware MixEd ReAlity (CAMERA) consists of $300k$ images with real background and rendered object instances from $1,085$ objects, of which $25k$ images and $184$ object instances are for validation. 
REAL provides $4,300$ training, $950$ validation and $2,750$ test images, with six categories of three objects for training and three for testing. For the validation set, another instance per object category is provided.
The challenge for these datasets is category-level pose estimation in multi-object scenes, with symmetries and occlusion in clutter.

\subsubsection{PhoCal~\cite{wang2022phocal}} The dataset shows $60$ specular-reflective, opaque and transparent objects, across eight categories. 
The scenes are multi-object and introduce occlusion.
The test set also provides novel instances per category.

\subsubsection{DREDS~\cite{dai2022dreds}} provides richly randomized synthetic training data of $119,580$ images, of which $19,380$ are for testing. 
These are composed of $1,801$ objects of seven categories.
An additional synthetic test set with $11,520$ images of $60$ novel objects is provided.
In the synthetic sets, object materials are randomized, ranging from diffuse over specular reflectivity, to transparency.
The real sets consists of $27k$ images of $42$ object instances from seven categories.
Additionally, $11k$ images of eight novel objects of known categories are provided.

\subsubsection{KeyPose~\cite{liu2020keypose}} $15$ object instances of five categories are provided.
The test sets show isolated object instances on diverse backgrounds.
The challenge is to detect and estimate poses of fully transparent objects.

\subsubsection{TransCG~\cite{fang2022transcg}} This grasping dataset provides $57,715$ images of $51$ transparent objects, of which $12$ are used for testing.
The challenge is objects' transparency, interacting with diverse opaque distractors and with changing background.
Opaque tags are attached to each object and present in the images.
One aspect that thus needs to be evaluated when using the dataset for training networks, is the influence of these tags on the learned feature encoding.

\subsubsection{HouseCat6D~\cite{jung2022housecat6d}} 
RGB, depth and polarimetric images are provided in $41$ scenes with $194$ object instances for pose estimation and grasping.
For pose estimation, $20k$ training, $3k$ test, and $1.4k$ validation images are available. 
These result in $160k$ poses, with unseen object instances of the $10$ categories in the test and validation set.
A total of 16 scenes have been annotated with $10M$ grasp points, which are intended for testing on a robot. 
An official split for evaluating grasping on the dataset has been established. 
The test and validation sets present challenges in the form of unseen object instances belonging to known categories, exhibiting strong reflections, lack of texture, and translucency.

\subsubsection{GraspNet/SuctionNet-1billion~\cite{fang2020graspnet,Cao2021suction}} Provided are $97k$ RGB-D images in $190$ cluttered scenes, with $10$ of $83$ objects per scene.
The dataset is split in $100$ scenes for training and $90$ for testing.
The test set is split in $30$ scenes of known instance, $30$ scenes of known categories, and $30$ scenes of completely novel objects.
These datasets provide $6D$ poses and grasp points for all objects.
Connected challenges are multi-object learning, due to the high number of object classes, and estimating poses of novel objects~\cite{gou2022unseen}. 

\subsection{Ongoing Research Problems}
\label{sec:trends}

This section presents the important current research problems listed in Table~\ref{tab:topics} in detail and illustrates their relevance for robotics. 
Works that fundamentally contribute solving the respective problems are presented and evaluated as to identify remaining open research questions.
Note that the order of presentation does not follow the order of frequencies of the respective motivating topics, in order to provide a more natural reasoning through the grouping of related open problems.

\subsubsection{Domain Shift}

Overcoming the domain shift between training and test data is an ever-present topic in machine learning. This problem is even more relevant in applied robotics, where gathering real data is expensive (e.g., due to robot operation or human demonstration) and the generation of synthetic training data thus is a compelling alternative.
Addressing this challenge in robotic perception, substantial progress has been made with respect to monocular object pose estimation. Various such strategies to overcome domain shift depend on the available data and annotations~\cite{li2020robust,thalhammer2021pyrapose,zhang2021keypoint,hu2022perspective,nguyen2022templates,wang2020self6d,shi2021fastUQ,lu2022slam,ikeda2022sim2real}.
At inference time, uncertainty-quantification improves robotic task execution under domain shift~\cite{shi2021fastUQ}, and online self-training strategies improve pose estimation during robotic execution~\cite{gu2022ossid,lu2022slam}.
The authors of~\cite{rad2018domain,zhang2021keypoint,li2020robust} align predictions in source and target domain without requiring annotation in the latter.
In~\cite{wang2020self6d,hu2022perspective,yang2022image}, the authors present solutions for effective synthetic-to-real transfer, where the addition of a few annotated samples in the target domain allows for even more effective alignment. 
Mimicking the appearance and noise patterns of the target domain using style transfer and few-sample learning allows to minimize the domain gap to the synthetic data~\cite{ikeda2022sim2real,yang2022image,Chen_2023_CVPR}.
Besides these strategies, especially domain randomization provides great potential for overcoming the sim-to-real gap of neural networks for robotic applications~\cite{tobin2017domain,sundermeyer2018implicit,li2019cdpn,zakharov2019dpod,thalhammer2021pyrapose}. 
Broad applicability in different target domains is ensured by assuming no data or annotation apart from that of the source domain.
Especially, employing Physically-based Rendering (PBR) enables rendering physically plausible reflections of materials, and shadows, which has been shown to be important for pose estimation~\cite{hodan2019photorealistic,zakharov2019dpod}.

Table~\ref{tab:bop_rgb} summarizes the Average Recall (\textit{AR}) for pose estimation, and the mean Average Precision (\textit{mAP}) for object detection, comparing methods that only use data from PBR for training and those that use a combination including real data.
The question arises if the difference in pose estimation performance when training on a combination of PBR and real data is insignificant and the observed performance improvements actually occur solely because more training data is used.
For TUD-L, the performance gains are higher, since more than eight times as many real samples are available per object, as compared to T-LESS.
For YCB-V, even more improvement is shown, which correlates with the increased amount of more than $18$ times as many real training images per object class.
Interestingly, the increase in detection performance is similar to that of pose estimation. 
The improvement in pose estimation correlates with the increased amount of data and is also partially attributable to better detection rates. %

\begin{table}
\caption{\textbf{BOP-Challenge results} Comparison of selected results with respect to used training data, taken from~\cite{hodan2022bop}. The Average Recall (\textit{AR}) is reported for pose estimation and the mean Average Precision (\textit{mAP}) is reported for object detection.} 
\centering
\begin{tabular}{|c|c|c|c|c|}
\hline
 Method & Data & T-LESS & TUD-L & YCB-V \\ \hline
 \multicolumn{5}{|c|}{Pose Estimation} \\ \hline
 GDRNet++\cite{liu2022gdrnpp_bop} & PBR & 79.6 & 75.2 & 71.3 \\ 
 GDRNet++\cite{liu2022gdrnpp_bop} & PBR+real & 78.6 & 83.1 & 82.5 \\ 
\hline
 \multicolumn{2}{|c|}{Performance difference} & -0.1\% & +9.5\% & +13.6\%  \\ \hline
 PFA\cite{hu2022perspective} & PBR & 71.9 & 73.2 & 64.8 \\
 PFA\cite{hu2022perspective} & PBR+real & 77.8 & 83.9 & 80.6 \\\hline
 \multicolumn{2}{|c|}{Relative performance change} & +7.6\% & +12.8\% & +19.6\%  \\ \hline
  \multicolumn{5}{|c|}{Detection} \\ \hline
 GDRNet++\cite{liu2022gdrnpp_bop} & PBR & 86.5 & 72.8 & 78.6 \\
 GDRNet++\cite{liu2022gdrnpp_bop} & PBR+real & 87.6 & 89.5 & 85.2 \\ \hline
  \multicolumn{2}{|c|}{Performance difference} & +1.3\% & +18.7\% & +7.7\%  \\ \hline
 PFA\cite{hu2022perspective} & PBR & 73.4 & 66.3 & 73.5 \\
 PFA\cite{hu2022perspective} & PBR+real & 79.8 & 86.6 & 85.0 \\\hline
   \multicolumn{2}{|c|}{Performance difference} & +8.0\% & +11.9\% & +13.5\%  \\ \hline
\end{tabular}
\label{tab:bop_rgb}
   \vspace{-1ex}
\end{table}

Table~\ref{tab:bop_paper} reports results of the BOP \cite{hodan2022bop} winning method GDRNet++ in comparison to the results reported in the supplementary material of GDRNet~\cite{wang2021gdr}. 
The performance difference is achieved by using stronger domain randomization, using ConvNeXt~\cite{liu2022convnet} as backbone instead of ResNet-34~\cite{he2016deep}, two mask heads for amodal and visible mask prediction, and by optimizing hyperparameters of the training process. 
This indicates that some performance improvement may already be gained through tuning existing methods, without providing sophisticated solutions for tackling the domain gap.

\begin{table}
\caption{\textbf{GDRnet~\cite{wang2021gdr}} Performance improvement achieved through hyperparameter tuning.}
\centering
\begin{tabular}{|c|c|c|}
\hline
 Method & LM-O & YCB-V \\ \hline
 GDRNet\cite{wang2021gdr} & 67.2 & 75.5 \\ 
 GDRNet++\cite{liu2022gdrnpp_bop} & 71.3 & 82.5 \\\hline
\end{tabular}
\label{tab:bop_paper}
   \vspace{-1ex}
\end{table}

Similar behaviour is observed in the VisDA-2021 challenge~\cite{visda2021}, that provides a benchmark for domain adaptation.
Training is performed on ImageNet1k~\cite{russakovsky2015imagenet}.
Both a development set with and a test set without annotations are provided. 
The development set consists of images and annotations from ImageNet-C~\cite{hendrycks2019imagenet}, ImageNet-R~\cite{hendrycks2020imagenet} and ObjectNet~\cite{barbu2019objectnet}, and introduces novel classes. 
For the test set no annotations are available and the class distribution varies from the development set.
For submitting the final model, it is allowed to use the development set for hyperparameter tuning, yet the test set can only be used for domain adaptation.

Table~\ref{tab:visda} reports the results of the top three methods using the classification ACCuracy (ACC) and the Area Under the Receiver Operating characteristic Curve (AUROC)~\cite{hand2001auroc}. 
The ACC quantifies the true positive rate in class prediction.
The AUROC quantifies the separation between known and unknown classes.
The two best performing methods use no domain adaptation strategies. 
The winning method achieves its performance through pretraining a transformer with ImageNet1k.
The second best method uses EfficientNetB7~\cite{tan2019efficientnet} in conjunction with extensive data augmentation and regularization.
These results indicate that tuning hyperparameters, such as data augmentations plays a predominant role for overcoming the domain shift.

\begin{table}[t!]
\caption{\textbf{VisDA-2021 challenge~\cite{visda2021}.} Results of the top three methods.}
\centering
\begin{tabular}{|c|c|c|c|}
\hline
 Ranking & Method & ACC & AUROC \\ \hline
 1 & Tayyab\tablefootnote{https://ai.bu.edu/visda-2021/assets/pdf/Burhan\_Report.pdf} & 56.29 & 69.79 \\ \hline
 2 & Rajagopalan\tablefootnote{https://ai.bu.edu/visda-2021/assets/pdf/Chandramouli\_Report.pdf} & 48.49 & 76.86 \\ \hline
 3 & Liao~\cite{liao20212nd} & 48.49 & 70.8 \\ \hline
\end{tabular}
\label{tab:visda}
   \vspace{-1ex}
\end{table}

Considering the leaderboards of the BOP and VisDA-2021 challenges, it appears that domain shift has been alleviated by using deeper models, extensive data augmentation, and further regularization techniques during training.
The presented considerations are only valid for standard datasets, where the variance of the data distribution in terms of, for example, object shapes, support planes, textures and illumination is limited.
Recent application-driven works consider significantly larger data-distribution shifts~\cite{hu2021wide,viviers2024advancing,sapienza2023underwater,tang2024rov6d,ulmer2024orbital}.
For example, different from the commonly studied industrial and home environments, robot perception for underwater and space applications~\cite{hu2021wide,ulmer2024orbital,sapienza2023underwater,tang2024rov6d} needs to deal with sensor noise that is vastly different due to optics, sensing principles and illumination. 
Driven by robotics-enabled exploration of such environments, these domains will naturally gain in relevance in the future.
Similarly, application-specific data domains like medical imaging will gain relevance due to the rising degree of automation and robotization of everyday tasks~\cite{viviers2024advancing}.

\subsubsection{Occlusion Handling}

\begin{figure}[t!]
   \centering
    \includegraphics[width=1.0\columnwidth]{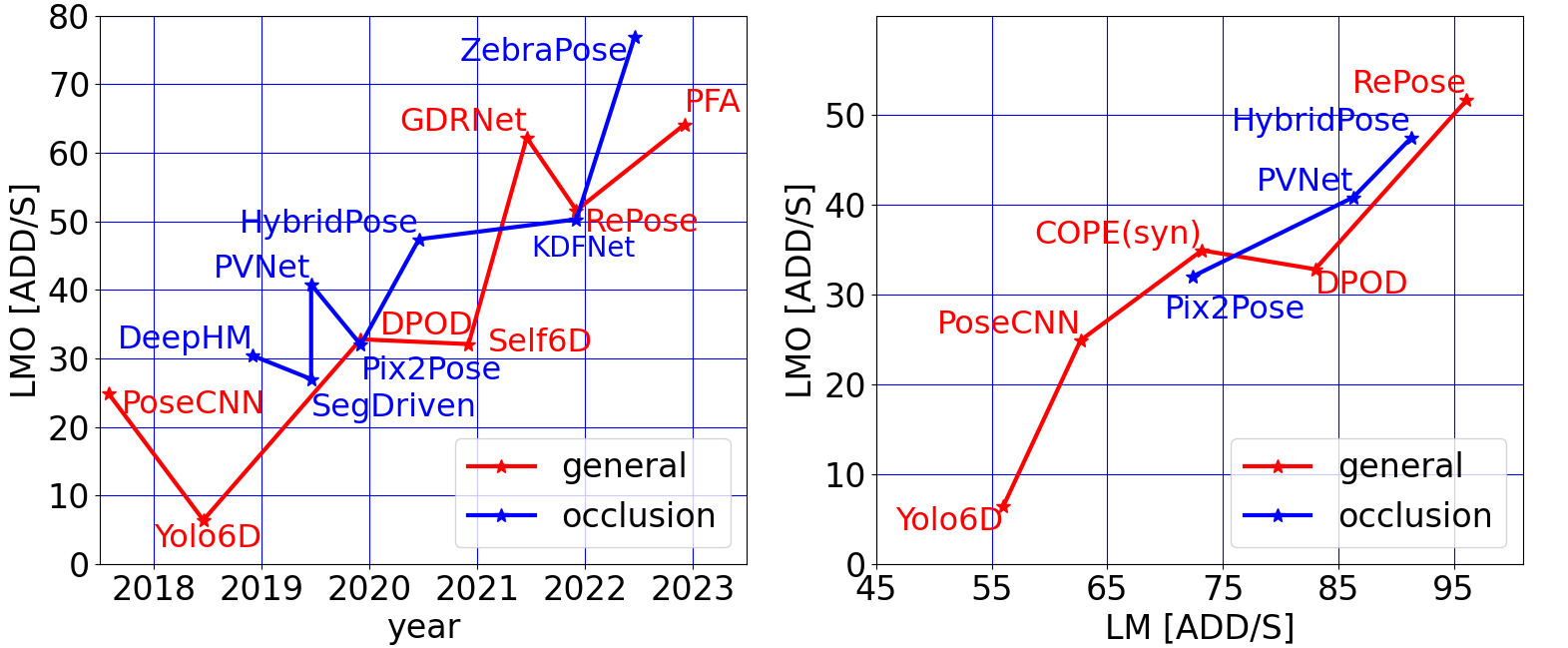}
   \caption{\textbf{Occlusion-handling performance.} Comparison of methods designed for occlusion handling (blue) and for general purpose (red). ADD/S used as evaluation metric~\cite{hinterstoisser2012model}.}
   \label{fig:occlusionProblem}
      \vspace{-2ex}
\end{figure}

Occlusion handling is an important challenge for object pose estimation and present in many real-world robotics scenarios. Dense clutter, a hand or gripper manipulating an object, or even an ill-selected viewpoint may result in large parts of the object of interest being occluded.
Table~\ref{tab:topics} shows that occlusion handling is one of the major challenges based on the frequency at which it is a motivating current research. 
Figure~\ref{fig:occlusionProblem} presents the performance of general purpose methods~\cite{hu2019segpose,oberweger2018making,peng2019pvnet,Park_2019_ICCV,song2020hybridpose,su2022zebrapose,zhang2021keypoint} compared to those designed for occlusion handling~\cite{hu2022perspective,iwase2021repose,tekin2018real,thalhammer2022cope,wang2020self6d,wang2021gdr,xiang2017posecnn,zakharov2019dpod}.
Evaluations on the LM~\cite{hinterstoisser2012model} and the LM-O~\cite{brachmann2014learning} dataset are presented.
LM features individual test sets for each of the $13$ dataset objects with no occlusion, while in LM-O eight LM objects appear under strong occlusion.
The left plot of Figure~\ref{fig:occlusionProblem} presents the performance of diverse methods on the LM-O~\cite{brachmann2014learning} dataset, plotted against the year of publication.
The performance of general purpose methods and those motivated by occlusion improved similarly over the years.
The right plot shows the correlation between the performance on LM-O and LM~\cite{hinterstoisser2012model}, for general purpose methods and those motivated by occlusion handling.
Both types of methods exhibit similar performance ratios.

The presented comparisons indicate that general pose estimation performance and occlusion handling correlate, since methods which handle occlusion well perform better overall.
Consequently, occlusion handling is not a specific trait to improve, but rather one that general pose estimation converges to.
A further indicator that reinforces this observation is that recently proposed datasets (later than $2017$) almost exclusively feature images with object occlusion (see Table~\ref{tab:datasets}).
The only exceptions are TUD-L, which focuses on challenging illumination changes, and KeyPose, focusing on transparency.

The substantial progress that has lately been made for handling occlusions is to be attributed to different influencing factors. 
Improvements with respect to data rendering not only reduced the domain shift between the rendered and the real domain, but also improved occlusion handling.
The major advantage of using rendered training data is that the distribution of aspects such as occluding patterns, illumination, and viewpoints alike, are arbitrarily diverse and thus reduce biases of trained estimators~\cite{homebrewedDB}.
Occlusion handling improvements are furthermore attributable to designing approaches that incorporate probabilities from local hypotheses~\cite{peng2019pvnet}, strategies for deriving poses from multiple pose representations~\cite{song2020hybridpose}, and to reasoning about self-occlusion.
However, while works exist that systematically analyse the influence of pose continuity~\cite{zhou2019continuity} and strategies for symmetry handling~\cite{richter2021handling}, occlusion handling research lacks comparable thoroughness.
Detailed systematic investigations of the influence of specific occlusion patterns, and the visibility of specific object parts, on the pose estimation accuracy are missing.
Yet, such investigations are required to design approaches that exhibit robust occlusion handling, and to provide reliable uncertainty estimates.  

\subsubsection{Pose representations}
\label{sec:poserep}

Many robotic tasks are facilitated by explicit geometrical information about the three-dimensional scene they need to, e.g., navigate or manipulate. For scenes decomposable into objects, shape and pose are used to represent this information. 
Parameterization is thus divided into the problem of estimating the values for the three translational, and the three rotational Degrees of Freedom (DoF) in the Euclidean space.
Parameterizing the $3DoF$ for translation requires strategies for handling the commonly used crop-level input~\cite{li2019cdpn}.
Estimating the $3DoF$ for rotation is challenging, due to object symmetries and rotation space ambiguities arising from estimating poses from a $2D$ space.
Deep learning pose estimation research explores Euler angels~\cite{sock2018multi}, quaternions~\cite{manhardt2018deep,xiang2017posecnn} and continuous $6D$ representations~\cite{zhou2019continuity}.
An unambiguous mapping between the object in the image space and its corresponding rotation values is crucial to effectively solve for object poses. 
Defining the $3DoF$ object rotation as the camera in object space, the allocentric rotation, alleviates such ambiguities~\cite{kundu20183d,wang2021gdr}.
Recent works adopt the indirect estimation of the rotation using template matching\cite{wohlhart2015learning,labbe2022megapose,nguyen2023nope}.

Early deep learning pose estimation works identified performance improvements when using keypoints as regression target instead of directly regressing the $6D$ pose~\cite{tekin2018real,crivellaro2015novel,rad2017bb8,xiang2017posecnn}.
Such geometric correspondences are nowadays the most commonly used surrogate training targets for representing $6D$ object poses~\cite{di2021so,haugaard2021surfemb,hodan2020epos,huang2022neural,hu2022perspective,su2022zebrapose,thalhammer2022cope,liu2021kdfnet}.
The $6D$ pose is derived by registering the estimated $2D$ correspondences to the corresponding ground-truth $3D$ ones.
Algorithms for such $2D$-$3D$ correspondence registration are either classical ones, like the Perspective-\textit{n}-Points (P\textit{n}P) algorithm~\cite{lepetit2009epnp}, or learned functions~\cite{chen2022epro,hu2020single,thalhammer2022cope,wang2021gdr}. 

\begin{figure}[t!]
   \centering
    \includegraphics[width=1.0\columnwidth]{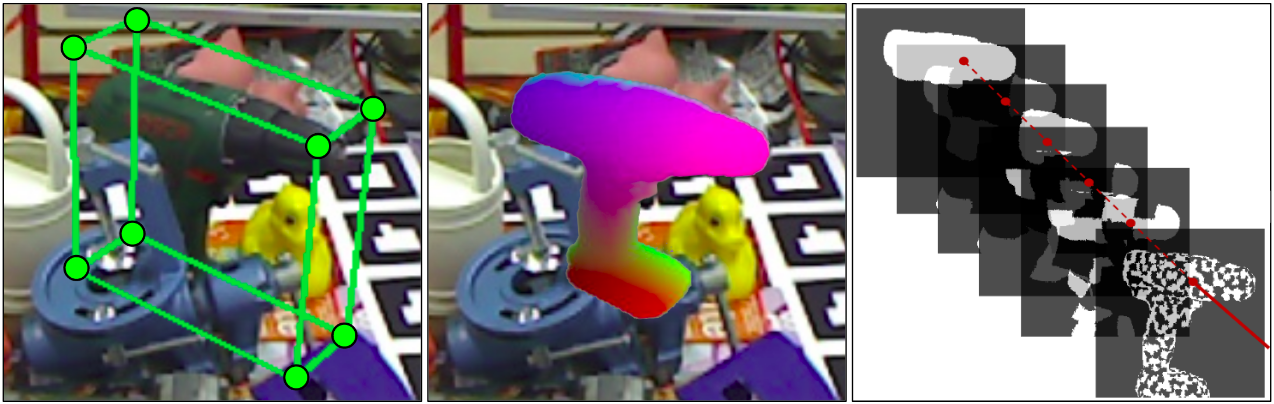}
   \caption{\textbf{Geometric correspondences.} Examples of sparse (left), dense (middle), and hierarchical dense geometric correspondences\protect\footnotemark (right).}
   \label{fig:gcs}
      \vspace{-2ex}
\end{figure}

The most commonly used surrogate training targets are either dense uv-coordinates~\cite{li2019cdpn,Park_2019_ICCV,zakharov2019dpod,huang2022neural,di2021so,liu2021kdfnet,hodan2020epos,haugaard2021surfemb,hu2022perspective} or sparse keypoints~\cite{crivellaro2015novel,rad2017bb8,peng2019pvnet,hu2019segpose,thalhammer2021pyrapose,liu2021kdfnet,zhang2021keypoint}.
For the sparse keypoints, improvements are proposed by assigning keypoints to geometrically relevant positions on the object surface and through more sophisticated keypoint location voting schemes~\cite{crivellaro2015novel,peng2019pvnet,liu2021kdfnet}.
Keypoints are also advantageous for guiding domain adaptation through aligning keypoint distributions across different domains~\cite{zhang2021keypoint}. 
Uv-coordinates are used as dense geometric correspondences for pose estimation~\cite{Park_2019_ICCV,li2019cdpn}.
It has been demonstrated that they are more versatile, e.g. with respect to symmetry handling~\cite{hodan2020epos}. 
Due to the dense prediction space, occlusion handling is improved~\cite{di2021so} and object symmetries can be learned in self-supervised ways~\cite{hodan2020epos,haugaard2021surfemb}.
Recently,~\cite{su2022zebrapose} proposed pixel-wise regression of binary patterns assigned to multiple hierarchical vertex groups, which also has been improved by~\cite{lian2023checkerpose}.
These hierarchical dense geometric correspondences lead to improved occlusion handling compared to their alternatives.
Figure~\ref{fig:gcs} presents visual examples of the aforementioned types of geometric correspondences.
Approaches that derive the pose from multiple representations are excellent candidates for improving robustness in robotics. This is because they overcome the shortcomings of individual representations and enable the quantification of uncertainties by registering consistencies of predictions made from different pose representations~\cite{song2020hybridpose,shi2021fastUQ,yang2023exploring}. 

\footnotetext{Courtesy of~\cite{su2022zebrapose}}

Currently proposed pose representations allow to improve occlusion and symmetry handling, domain adaptation and pose estimation accuracy in general. We conjecture that investigating novel representations to be a viable direction to significantly improve performance in both existing and currently unsolved tasks.
Future directions (see Section~\ref{sec:future}) will be providing improved solutions for category-level pose estimation, challenging material properties, and deformable and articulated objects.
As an alternative, learning principles that do not explicitly use $6D$ poses or $2D$-$3D$ correspondences as regression targets is gaining momentum~\cite{nguyen2022templates,wang2020self6d,Chen_2023_CVPR,guo2023knowledge,thalhammer2023self}.
Self-supervision possesses the potential to improve over manually designed pose representations, since it allows learning semantic correspondences between different objects and domains~\cite{florence2018dense}.
Recently it has been shown that directly deriving $6D$ poses from geometric correspondences in an end-to-end trainable manner improves pose estimation accuracy over only regressing geometric correspondences and deriving the pose using P\textit{n}P~\cite{hu2020single}.

\subsubsection{Multi-object and End-to-end Training}
\label{sec:e2e}

Handling multiple different objects is important because robotic scenarios, like assistant tasks in households, feature a multitude of objects with diverse properties.
The concept of multi-object pose estimation is strongly interleaved with end-to-end trainability, yet a clear differentiation is missing.
For the sake of clarification, we hereby break down these concepts.
By relating them to the respective concepts for object detection highlights the differences, and thus, uncovers potential challenges and aspects for improving pose estimation.

The state of the art for object detection groups algorithms into single-staged~\cite{lin2017focal,bochkovskiy2020yolov4,ge2021yolox,tian_fcos} and multi-staged ones~\cite{he2017mask,zhou2021probabilistic}.
The difference between these two types is depending on the usage of intermediate object proposals.
Single-staged object detectors extract features and predict attributes such as object class and bounding boxes directly.
Multi-staged approaches extract features, create a set of object proposals, which are in turn used for predicting the desired attributes.
Object proposals often already contain class, location and scale information~\cite{zhou2021probabilistic}.
Both types of approaches are single networks that handle a) multiple object classes and b) multiple instances simultaneously, and c) provide the desired object attributes directly, without requiring additional separately trained or executed stages.
The predicted object attributes are $2D$ bounding box coordinates for object detection.

Object pose estimation research treats these concepts differently.
Single-stage refers to detecting and estimating poses in the same end-to-end trainable stage~\cite{hu2020single,thalhammer2022cope,lin2022single}. 
Yet, this does not necessarily mean that multiple instances of the same object are handled~\cite{hu2020single}.
End-to-end trainability in object pose estimation research refers to directly estimating the $6D$ pose and backpropagating the respective loss for the pose to layers that learn features, or regress geometric correspondences~\cite{di2021so,thalhammer2022cope,wang2021gdr}.
Yet, this end-to-end trainability is not interchangeable with the concept of single-stage approaches, as is the case for object detection.
Such approaches are multi-staged, requiring a detector for estimating sparse location priors~\cite{di2021so,wang2021gdr}.
The following paragraphs presents those two concepts, multi-object and end-to-end trainable pose estimation, in detail and presents the connected challenges.

\textbf{Multi-object Approaches:}
The main problems connected to multi-object learning is that imbalances in the training data need to be handled to reduce the network bias.
These biases reduce estimation accuracy of objects underrepresented in the training data and typically are training samples per object, different object scales and different convergence behaviour per object, caused by aspects such as texture, geometry and material properties. 
Yet, the advantages of multi-object training are abundant and unanimously important for robotic applications.
Recent approaches present advantages with respect to runtime, scalability, memory footprint and general applicability~\cite{gard2022casapose,hodan2020epos,thalhammer2022cope,thalhammer2021pyrapose,zakharov2019dpod, viviers2024advancing,chaitanya2022physics}.
Despite these obvious advantages, the performance trails behind single-object approaches, which train separate pose estimators per object.

\textbf{End-to-end Trainable Approaches:}
Recently, advances have been made by additionally supervising learning with the direct $6D$ pose as downstream learning task.
As such, geometric correspondences are regressed as an intermediate representation and the $6D$ pose is inferred from them.
Training in such an end-to-end manner improves performance over regressing the 6D pose directly, or using P\textit{n}P and alike~\cite{hu2020single,wang2021gdr,thalhammer2022cope,iwase2021repose,park2022dprost,cao2022dgecn,liu2023linear}.

Direct pose regression also provides increased diversity for supervision~\cite{di2021so,thalhammer2022cope}.
By using the estimated $6D$ pose for re-projecting the geometric correspondences, keypoints~\cite{thalhammer2022cope} or uv-coordinates and self-occlusion maps~\cite{di2021so}, training undergoes additional supervision.
End-to-end training of pose estimators provides means to enforce consistency with tasks learned in parallel.
As such encoded representations are more general and do improve through multi-task learning~\cite{lopes2023cross}.

\subsubsection{Refinement}

Classical multi-stage depth-based approaches~\cite{hinterstoisser2012model,drost2010model,vidal20186d,hodavn2015detection,aldoma2011cad} refine estimates using ICP~\cite{rusinkiewicz2001efficient}.
Also methods that estimate poses in a monocular fashion~\cite{kehl2017ssd,Park_2019_ICCV,labbe2020cosypose,xiang2017posecnn} exploit additional depth information for ICP-based refinement.
To be able to rely only on monocular images thus requires alternative solutions for pose refinement.
Some early learning-based works use contour-based~\cite{kehl2017ssd,manhardt2018deep} or keypoint-based refinement~\cite{rad2017bb8}.
A seminal work that has been adopted and modified often is presented in~\cite{li2018deepim}.
An input image pair, the observation and a rendered estimate, is processed by a network to predict their relative transformation.
Many recent works build on this iterative refinement approach~\cite{labbe2020cosypose,hu2022perspective,lipson2022coupled,zakharov2019dpod,iwase2021repose,haugaard2021surfemb,araki2021iterative,yen2021inerf}.
Recent works propose to replace iterative refinement by parallel hypotheses scoring and matching against templates~\cite{hu2022perspective,he2022generative}.
Yet, these methods still benefit from using an additional iterative refinement stage~\cite{hodan2022bop}.
A common aspect of these approach is their render-and-compare nature, creating a dependence on (textured) 3D object models which is not easily address for, e.g., novel objects. Furthermore, these approaches need to be able to render the object of interest with reasonable realism. However, for complex materials found in highly reflective or refractive objects, rendering may be prohibitively compute intensive. Also, this typically relies on material properties that are not readily available for existing models and are difficult to acquire for novel objects.

This is an especially limiting requirement as, lately, a considerable shift is happening toward pose estimation for objects with more challenging material properties like metallic and transparent materials~\cite{liu2020keypose,wang2022phocal}.
Environmental properties, like illumination and light direction, and scene background have to be inferred in order to synthesize templates accurately enough to precisely match observations against them.

Generally, the question of the need for pose refinement arises since single-shot pose estimation improves rapidly, and use cases such as object grasping tolerate certain pose inaccuracies.
Similarly, robotic grippers are designed to tolerate grasp pose inaccuracies and as such often execute successfully, even when the estimated pose is visually significantly off.
Yet, since robotic scenarios like 
affordance-based grasping requires higher accuracy, we conjecture that pose refinement will remain relevant in the near future.

\subsubsection{Symmetry Handling}

Object symmetries are comparably easy to handle for classical template-based approaches.
Templates are rendered in a viewing sphere around the object model. Views are encoded into descriptors to create a look-up table of the training samples~\cite{drost2010model,hodavn2015detection,aldoma2011cad,hinterstoisser2012model}.
Such template matching approaches do not require learning a representation or regressing a pose.
Matched templates with ambiguous views, but incorrect poses, are thus handled by common metrics~\cite{hinterstoisser2012model,rad2017bb8,hodan2020bop}.
Yet, incorporating information about symmetries still helps reducing the memory footprint and runtime~\cite{alexandrov2019leveraging}.

Deep learning approaches, in contrast, encode representations of the training data.
Approaches based on supervised learning, which are the most common ones, require the $6D$ pose representations for backpropagation during training.
These representations are often not invariant to visual object ambiguities.
Deep learning approaches thus suffer from pose estimation accuracy reduction when such visual ambiguities are not considered at training time~\cite{rad2017bb8,sundermeyer2018implicit,xiang2017posecnn}.

\begin{figure}[t!]
   \centering
    \includegraphics[width=0.8\columnwidth]{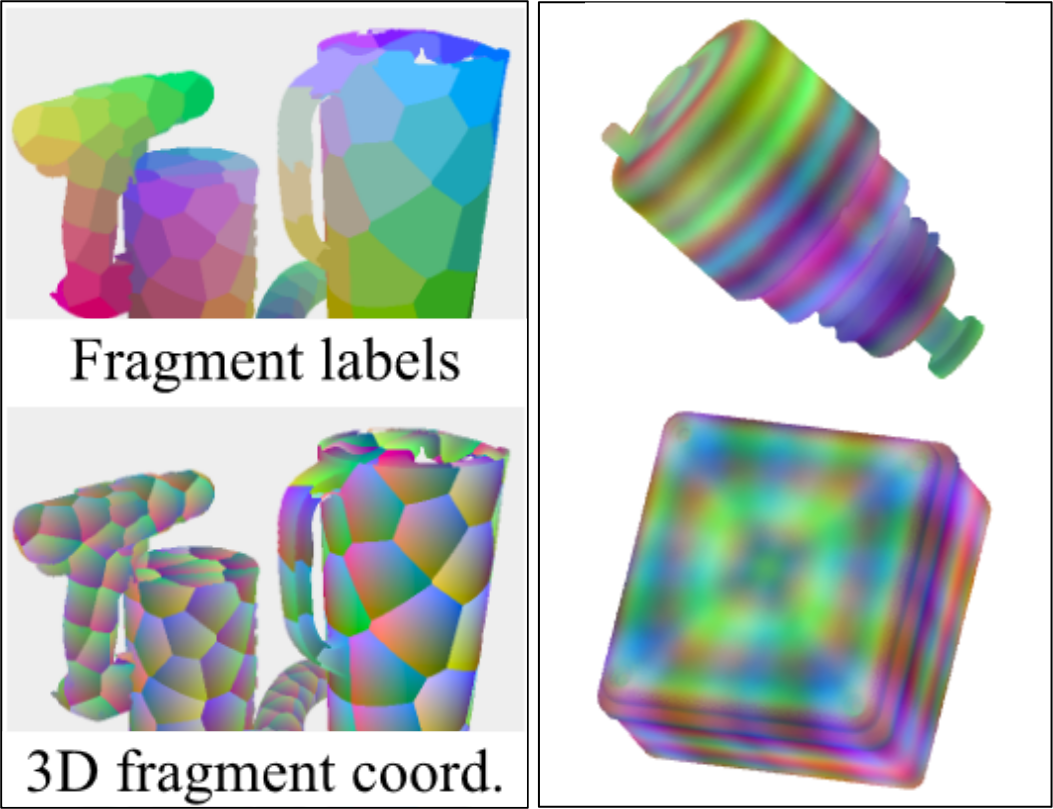}
   \caption{\textbf{Symmetry handling.} Surface fragment classification\protect\footnotemark~\cite{hodan2020epos} (left), surface correspondence embedding~\cite{haugaard2021surfemb} (right).}
   \label{fig:uv-sym}
      \vspace{-2ex}
\end{figure}

\footnotetext{Courtesy of~\cite{hodan2020epos}}

Common strategies to tackle this problem are limiting the range of views seen during training~\cite{rad2017bb8},
using loss functions that account for ambiguous views~\cite{xiang2017posecnn,Park_2019_ICCV,thalhammer2022cope}, and designing network structures to provide multiple hypotheses for disambiguation~\cite{manhardt2019explaining,shi2021stablepose}. 
For these approaches to work, oftentimes symmetries have to be known beforehand which limits their applicability and generality.  
Furthermore, manually assigning symmetries is also a potential source of error and standard geometric correspondence formulations are often insufficient~\cite{richter2021handling}.
When using uv-coordinates as regression targets, sophisticated solutions for solving object symmetries during training exist.
Figure~\ref{fig:uv-sym} shows a visualization of the approaches in ~\cite{hodan2020epos}, which learns symmetries by predicting ambiguous surface regions, and~\cite{haugaard2021surfemb}, which encodes geometric representations invariant to symmetries. 
More recent approaches also estimate ambiguities in the output space\cite{jeon2023ambiguity}.

An open problem is finding such solutions for keypoint-based approaches, whose sparseness is beneficial in terms of runtime and scalability as compared to uv-coordinate approaches.
Robotic applications may additionally benefit from finding semantically important object locations that enable direct grasp point estimation and affordance-based grasping using keypoint-based approaches~\cite{ju2024robo}.
In general, diverse strategies exist for handling symmetrical objects, and modern approaches experience minor drops in pose estimation accuracy when faced with them.

\subsubsection{Category-level Training}

Instance-level approaches aim at retrieving poses of known object instances. By comparison, category-level pose estimation aims at generalizing to unknown object instances of the same category~\cite{he2022fs6d,chen2020category,lin2022single,ma2022robust,fan2022object,wen2022disp6d,Lee2021category,remus2023i2c-net,deng2022icaps,di2024zero123,fan2024acr}.
Common principles are encoding canonical or category-specific features~\cite{lin2022single,chen2020category,fan2022object,wen2022disp6d} and render-and-compare~\cite{ma2022robust}. 
In~\cite{chen2020category}, category-specific features are used to reconstruct object views conditional on a pose. 
During inference, the network is iteratively optimized for pose and shape.
Alternatively, the authors of~\cite{Lee2021category} retrieve object poses by aligning predicted depth with object coordinates.
Similarly, in~\cite{lin2022single} the authors estimate the $6D$ pose by aligning predicted object depth and coordinate from an extracted shape prior.
In~\cite{wen2022disp6d}, latent representations are compared to codebook encodings for retrieving the object pose. 
The authors of~\cite{ma2022robust} employ a contrastive learning framework, from which object proposals are generated and compared against rendered templates, for retrieving the $6D$ pose.

Table~\ref{tab:category} compares approaches based on the input image modality on CAMERA25 and REAL275~\cite{wang2019normalized}.
We advise readers to take this as a qualitative comparison. 
First, the presented methods do not use the exactly same training and test data.
Secondly, some of the results may be faulty, as indicated by~\cite{you2022cppf}, who report updated NOCS results after fixing their erroneous $3D_{50}$ metric computation.
Nevertheless, the presented comparison shows that the pose and keypoint estimation accuracy for approaches with RGBD input increased significantly as compared to RGB-only approaches. 
The goal of this comparison is to highlight the open challenge of closing the huge performance gap observed for monocular approaches, which seems to be widening since RGB-only research stagnated between $2023$ and $2024$.

\begin{table}
\setlength{\tabcolsep}{5pt}
\caption{\textbf{Category-level Pose Estimation} Comparison of RGB and RGB-D input on the CAMERA25 and REAL275 datasets~\cite{wang2019normalized}.
Results taken from the original papers; empty entries were not provided. Note that these results are on inconsistent train-test splits. The numbers indicated with \textdagger are from~\cite{you2022cppf}.}
\centering

\begin{tabular}{|c|c|c|c|c|c|c|}
\hline
 Method & Year & Input & \multicolumn{2}{|c|}{CAMERA25} & \multicolumn{2}{|c|}{REAL275} \\ 
 \cline{4-7}
 & & & \multirow{2}{*}{$3D_{50}$} & \ang{10} & \multirow{2}{*}{$3D_{50}$} & \ang{10} \\
 & & & & 10 cm & & 10 cm \\\hline
 NOCS\cite{wang2019normalized} & 2019 & RGB\textbf{D} & - & 62.2 & 27.8$^{\dagger}$ & 26.7 \\
 Zheng et al.\cite{zheng2023hs} & 2023 & RGB\textbf{D} & 93.3 & - & 82.8 & - \\
 Remus et al.\cite{remus2023i2c-net} & 2023 & RGB\textbf{D} & - & - & 92.5 & 67.1 \\\hline
 Chen et al.\cite{chen2020category} & 2020 & RGB & - & - & - & 4.8 \\
 Lee et al.\cite{Lee2021category} & 2021 & RGB & 32.4 & 19.2 & 23.4 & 9.6 \\
 Fan et al.\cite{fan2022object} & 2022 & RGB & 32.1 & 23.4 & 25.4 & 9.8

 \\\hline
\end{tabular}
\label{tab:category}
\end{table}

\begin{figure}[t!]
   \centering
    \includegraphics[width=1.0\columnwidth]{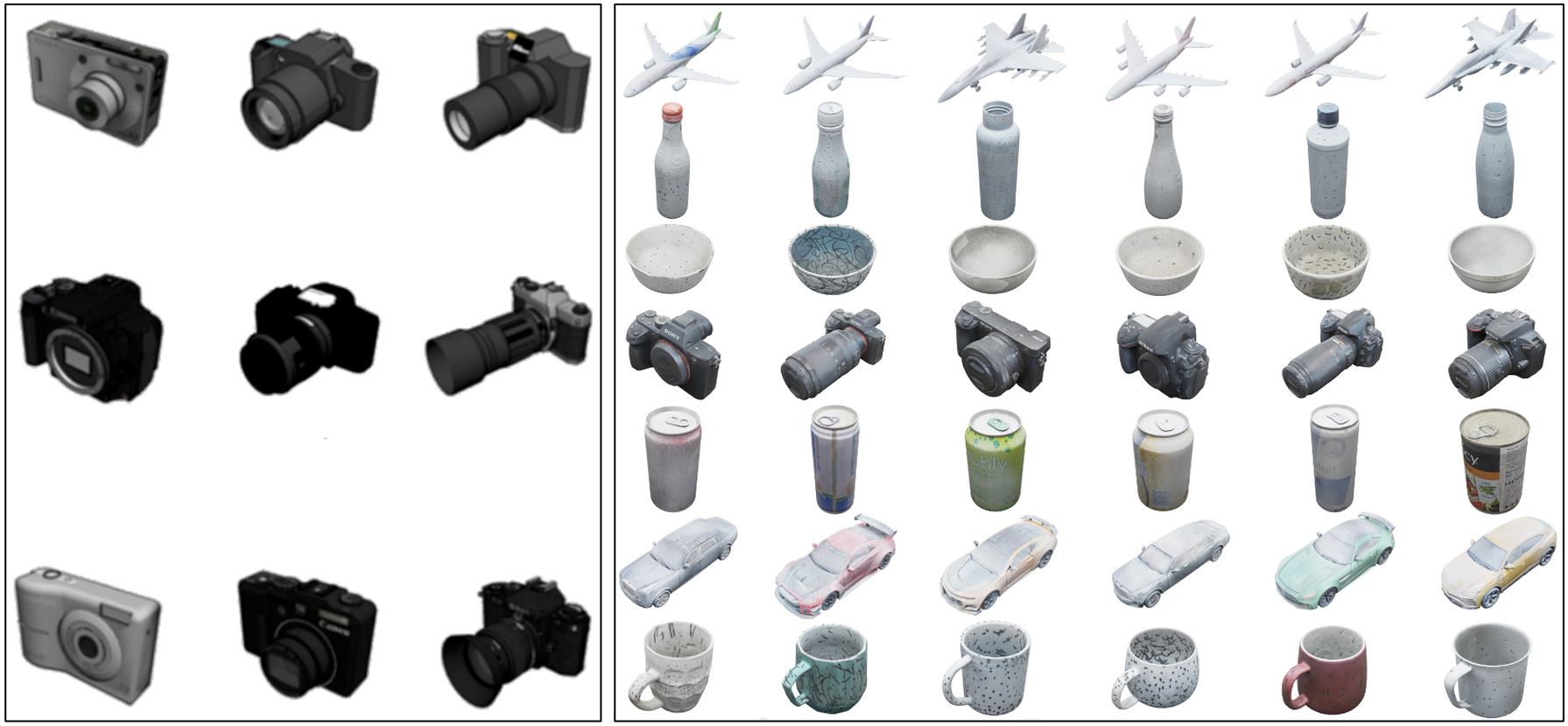}
   \caption{\textbf{Object-category variations.} Visualized are instances of the category \textit{camera} of NOCS, and the $42$ objects of DREDS with the categories along the rows\protect\footnotemark.}
   \label{fig:category_viz}
      \vspace{-2ex}
\end{figure}

Inspecting category-level pose estimation from a broader perspective, there are gaps that need to be filled in order to reap its full potential for unstructured open-world scenarios, where it is impractical to employ instance-level pose estimators for all objects.
The intra-category variation of the standard datasets used for designing and testing approaches is narrow.
Figure~\ref{fig:category_viz} shows the individual instances of the known categories of DREDS, and the category \textit{camera} of NOCS.
The categories of DREDS and NOCS are very similar and exhibit significant overlap.
Four of NOCS categories: \textit{bottle}, \textit{bowl}, \textit{camera}, \textit{can}, \textit{laptop}, and \textit{mug}, also appear in DREDS.
Both datasets have little intra-category, but large inter-category variation.

Important questions to answer are: Which variation of instances is required to effectively extrapolate to unknown instances of a category? 
How to handle objects that lie at the intersections of two similar categories? 
Which pose to retrieve in such cases? 
And how to circumvent these cases in the real world?
In order to answer these questions, formal rules for grouping object instances into categories are required. 
Future work thus has to investigate taxonomies of category-level pose estimation.
Solving this problem potentially requires algorithms that establish this taxonomy based on what instance-related features can be learned and as such category clustering itself.
Ideally, these also define the object origins to circumvent algorithms to interpolate the estimated pose between two categories with obviously diverging origins.

\footnotetext{Courtesy of~\cite{wang2019normalized} and~\cite{dai2022dreds}}

An interesting aspect of category-level approaches is that they generalize beyond known objects, within known categories, when compared to instance-level ones, yet still exhibit fast inference.
Alternatives, if fast inference is not a requirement, are novel object pose estimation approaches.

\subsubsection{Novel Objects}

Instance- and category-level approaches do not generalize to unknown object categories.
Novel object pose estimation aims to solve this task and doing so from monocular images is an emerging topic~\cite{gu2022ossid,li2018deepim,shugurov2022osop,nguyen2022templates,Sun_2022_onepose,goodwin2022zero,thalhammer2023self,okorn2021zephyr,wen2024foundationpose}.
Common strategies are using support views~\cite{liu2022gen6d,fan2023pope}, template-matching~\cite{nguyen2022templates,shugurov2022osop} from renderings, and foundational models~\cite{goodwin2022zero,thalhammer2023self,wen2024foundationpose}. 

Approaches retrieving the object pose using support views require labelled views around the object~\cite{liu2022gen6d,Sun_2022_onepose}.
These approaches assume the availability of real, but less, views as compared to approaches that use rendered templates.
Recently,~\cite{fan2023pope} relaxed these assumptions, only requiring a single image without annotation for retrieving a relative pose.
While the generality and the requirement of only one support view allows easy applicability, the pose estimation accuracy lags behind that of rendered template-based approaches, such as~\cite{nguyen2022templates,shugurov2022osop,thalhammer2023self}.
Such approaches match the query image against uniformly sampled views of the object to retrieve the query's pose.
This is done by computing the mutual similarities of the features extracted using fine-tuned CNNs.
An alternative is to use foundational models such as Vision Transformer (ViTs)~\cite{dosovitskiy2020image} pre-trained on ImageNet1k~\cite{goodwin2022zero} in a self-supervised manner~\cite{goodwin2022zero,thalhammer2023self,fan2023pope}.
Using pre-trained ViTs is promising to generalize to arbitrary objects, and circumvent the requirement for fine-tuning. 

Novel object pose estimation has great potential for example for assistance robots that encounter a vast number of objects over the course of their deployment. Using these approaches, estimating poses of such objects that are unknown during training does not require re-training the pose estimator~\cite{labbe2022megapose}.
The challenge of deep template matching for novel object pose estimation is also identified as an important task in the next edition of the BOP-challenge.
There is still much space for improvements with respect to accuracy when using RGB and efficiency, i.e. object properties, which are limited to opaque objects at the moment, and runtime, which is depending on the number of templates that are compared against the query.
As such it is expected that future research will solve these issues by exploring alternative ways to retrieve poses of novel objects without requiring rendered templates.
Furthermore, considering the current solutions and their limitations with respect to the object used, bridging novel object pose estimation to robotics, especially mobile robotics, will be challenging and important.
Considering unstructured open-world scenarios, novel object pose estimators need to deal with strong illumination changes, different cameras, differences between the available object models for creating templates and the query object, and different challenging material properties such as specularity and transparency.
Especially, the problem of handling challenging object material, is still largely unsolved even for monocular instance-level pose estimation.
Substantial progress has to be made to transfer the high pose estimation accuracy on opaque objects to novel objects with challenging material properties.

\subsubsection{Challenging Material Properties}
\label{sec:obj_material}

Most of the state of the art research primarily focuses on a limited range of object surface material properties. 
Considering the BOP, six of the seven core datasets feature fully opaque and diffusely reflecting objects~\cite{brachmann2014learning,homebrewedDB,hodan2017tless,doumanoglou2016recovering,xiang2017posecnn,hodan2018bop,drost2017introducing}. 
Four of those datasets contain primarily textureless objects~\cite{brachmann2014learning,homebrewedDB,hodan2017tless,drost2017introducing}.
Only one of these seven datasets includes specular, metallic object surfaces~\cite{drost2017introducing}.

\begin{figure}[t!]
   \centering
    \includegraphics[width=0.8\columnwidth]{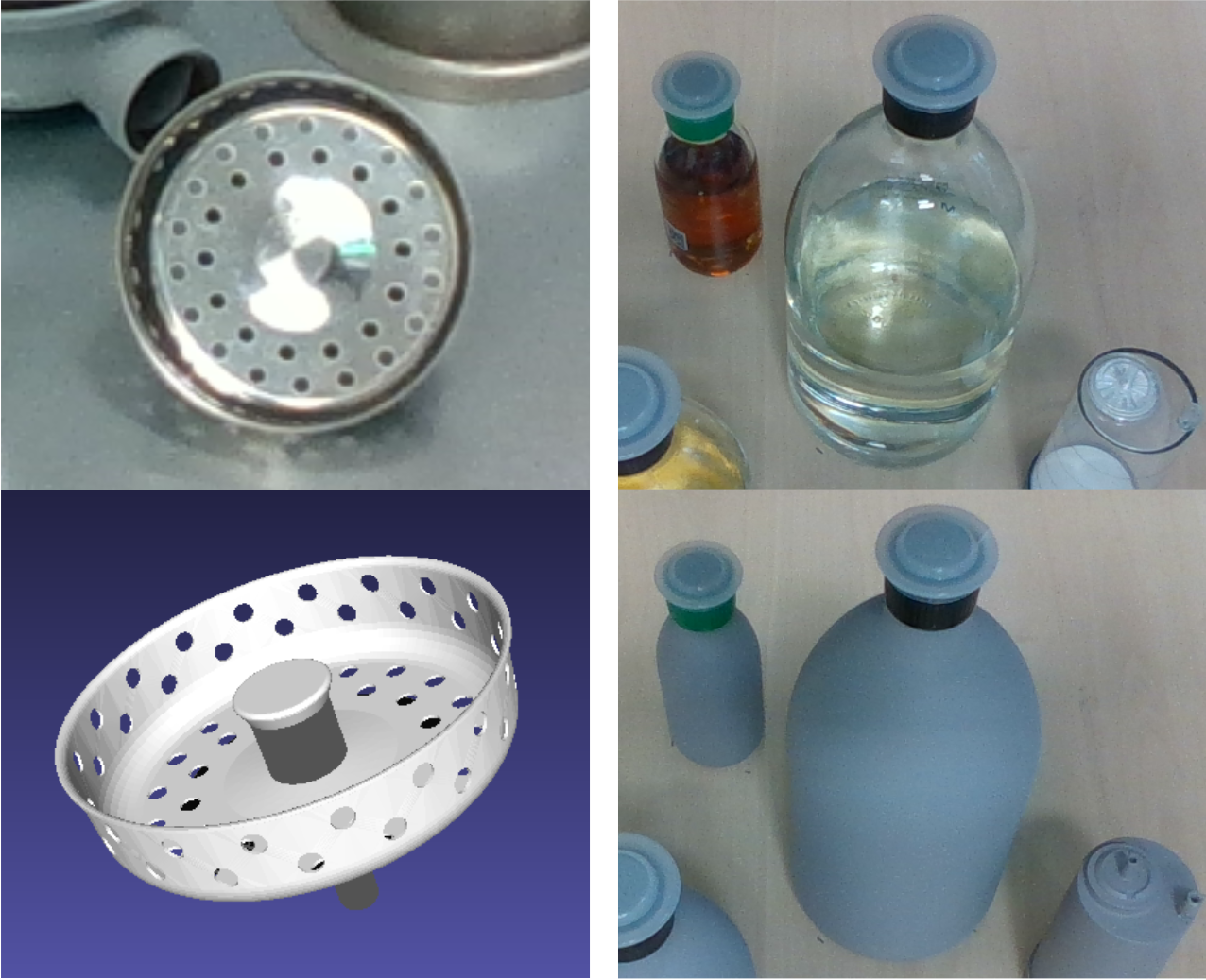}
   \caption{\textbf{Challenging object materials.} The left column compares the image of a drain strainer to its CAD model. The material's specular reflectivity makes it difficult to retrieve the object geometry from the captured image. The right column compares transparent objects to their opaque counterpart. Based on the viewpoint and filling, the transparent objects' appearance varies largely.}
   \label{fig:met_trans}
      \vspace{-2ex}
\end{figure}

Metallic surfaces are challenging due to their specular reflectivity, which leads to higher variation in appearance, depending on the incoming light and camera position, visualized by the left images of Figure~\ref{fig:met_trans}.
Especially for industrial application, handling metallic objects is of great importance.
Yet, solutions for handling them are comparably scarce as compared to diffusely reflective objects~\cite{he2022generative,drost2017introducing}.
The authors of~\cite{drost2017introducing} propose a metallic object dataset and solve pose estimation using classical methods on grayscale images.
In~\cite{he2022generative}, a non-publicly available dataset is used for pose estimation of specular reflective objects in clutter, from RGB. 
They match query images against templates, using geometric edge representations as input.
While this method provides a feasible solution for metallic object pose estimation and grasping, an empirical analysis of how much accuracy improvement is to be expected when using edge representation, and not RGB images, as input is currently missing. 
This may help to identify the specific scene propreties and situations where RGB input leads pose estimation accuracy to drop.
Having this information would on the one hand help to robustify algorithms against image artefacts occurring due to specular reflectivity of the object, and on the other hand, it would facilitate the design of benchmarking datasets that feature this specific challenges, deviating less from realistic scenarios.

Even more challenging are transparent object surfaces.
Instead of showing the texture of the object itself, RGB observations may only show a refracted view of the background that is depending on the transparent object, visualized by the right images of Figure~\ref{fig:met_trans}. Additionally, many transparent objects are also specular reflective.
Especially for robotics, transparent objects are of great importance because they are frequent in artificial environments, for example, glassware or food containers.
Handling such objects gained considerable momentum in computer vision~\cite{zhang2022transnet,chen2022clearpose,liu2020keypose,ichnowski2021dex,chen2022stereopose}.
Approaches for retrieving object poses rely on reconstructing depth data or on multi-view registration~\cite{chen2022clearpose,liu2020keypose,chen2022stereopose}.
Similar to the approach of~\cite{he2022generative} that showed the usefulness of edge images as input for metallic object pose estimation, it has been shown that such edge representations also improve pose estimation of transparent objects~\cite{yu2023TGFnet}. 
Concurrent research showed that, alternatively, extensive randomized synthetic data helps algorithms to encode meaningful visual cues for transparent object pose estimation~\cite{byambaa20226d}.
For future work, thorough investigation of the strengths and weaknesses of RGB and depth data for transparency handling is advisable, and will be crucial when considering the requirements of vision systems for robotics.

Finding viable solutions for challenging material properties requires bridging computer vision, concerned with analysis, and computer graphics, concerned with synthesis.
Approaches at this intersection do not only aim to analyze data, but also to encode physically-based functions for synthesizing such data~\cite{ichnowski2021dex}. 
This requires diverse learning principles and algorithm designs at that intersection, such as Neural Radiance Fields and Gaussian Splatting~\cite{mildenhall2021nerf,li2022nerf,ho2020denoising,kerbl20233d}.

\subsubsection{Beyond Supervised Learning}

Due to the complexity of deriving $6D$ poses from $2D$ input, object pose estimation is traditionally approached as a supervised learning task.
Yet, alternative training principles expose a number of advantages which also have been applied to pose estimation.

Self-supervised learning improves pose estimation accuracy, when data without annotation is available in the target domain~\cite{wang2020self6d,Chen_2023_CVPR,wang2021self6d++,gu2022ossid}.
The authors of~\cite{wang2020self6d,wang2021self6d++} show performance improvements using a limited number of real images without annotations, and differentiable rendering.
The approach of~\cite{gu2022ossid} enables pose accuracy improvement for novel objects using an online self-supervised learning scheme.
Using images of the object in the target domain and a coarse pose,~\cite{Chen_2023_CVPR} 
improves pose estimation while simultaneously improving the texture quality of the object meshes.
Contrastive learning not only shows great potential for novel object pose estimation in RGB~\cite{shugurov2022osop,nguyen2022templates,labbe2022megapose}, but also for knowledge distillation from large to small networks~\cite{guo2023knowledge}.
Such learning schemes, in combination with ViTs, are particularly interesting since they are general feature extractors and as such provide robust image-to-image correspondences without fine-tuning~\cite{caron2021emerging,thalhammer2023self,goodwin2022zero}.
Reinforcement learning (RL) approaches in pose refinement allow to limit the data annotation effort by exploiting segmentation masks as proxy for pose annotations~\cite{shao2020pfrl}, or by defining an expert policy that annotates new samples as the agent explores the pose space during training~\cite{bauer2021reagent}. Also, in the framework of RL, utilizing a replay buffer of refinement sequences allows to train with a large number of refinement steps. While end-to-end supervised refinement \cite{wang2019densefusion} needs to retain information about all steps for backpropagation and is hence memory limited, in RL temporal consistency is enforced by return computation. This allows to train on smaller batches from the replay buffer. 
Another benefit of RL is that any additional learning objective that is not differentiable may be incorporated in the reward~\cite{bauer2022sporeagent}. 
To improve the convergence of RL-based methods, they are often combined with Imitation Learning in the form of Behavioral Cloning~\cite{shao2020pfrl,bauer2021reagent,bauer2022sporeagent}.
Robot action modeling with RL also enables object singulation for simplifying object pose estimation in cluttered scenes~\cite{kiatos2019singulation,sarantopoulos2020singulation}.

Researching non-supervised learning principles encourages alternative research avenues for tackling, for example, domain adaptation, few-shot learning, overcoming the requirement for annotations, and knowledge distillation.
The success of self-supervised learning with vision transformers~\cite{caron2021emerging}, and in particular their feature encoding, also suggests possible performance improvements.

\subsubsection{Uncertainty Estimation}

Scores for estimating uncertainty are essential for virtually all tasks in computer vision, from detection scores~\cite{tian_fcos}, over descriptor similarities~\cite{sundermeyer2018implicit}, to inlier ratings~\cite{fischler1981random,lepetit2009epnp}.
Pose estimation approaches generally benefit from providing many hypotheses that are ranked by a confidence score.
On the one hand, evaluation metrics are primarily sensitive to accuracy~\cite{hodan2020bop}, and on the other hand robotic scenarios benefit from pruning hypotheses in order to retrieve the one with the highest confidence.
Uncertainty estimation allows to select the best object pose for use in down-stream tasks. When all pose hypotheses are uncertain, for example, it may be better for the robot to change its viewpoint rather than attempting a potentially unsuccessful manipulation. The notion of uncertainty may also be used to explain the robot's behavior in HRI scenarios. Finally, we may opt to spend additional computation time on refining the most promising hypotheses when the task allows it.
Ideally, confidences are retrieved from hypotheses distributions that describe actual probabilities.

Uncertainties are used in different ways to improve pose estimation accuracy.
In~\cite{peng2019pvnet}, a modified P\textit{n}P is proposed to incorporate the $2D$ distribution of predicted keypoint locations.
The authors of~\cite{manhardt2019explaining} identify an object's axis of ambiguity by analyzing the pose hypotheses distribution. 
In~\cite{deng2021pose}, object ambiguities are identified through deriving rotation estimation uncertainties from consecutive frames. 
The authors of~\cite{hu2019segpose} and~\cite{thalhammer2022cope} aim at deriving confidences per hypotheses, in order to retrieve a pose averaged over hypotheses with high accuracy.
Uncertainty estimation is achieved by quantifying the disagreement over an ensemble of pose estimators in~\cite{shi2021fastUQ}.
The authors of~\cite{huang2022confidence} learn to predict keypoint confidences in an unsupervised fashion.
In~\cite{jeon2023ambiguity}, object ambiguities are leveraged to derive uncertainties for keypoint selection.
Recently,~\cite{Yang_2023_CVPR} proposed a framework for estimating uncertainties from keypoint heatmaps that correlate with the estimation deviation from the ground truth keypoint locations.

\subsubsection{Applications for Monocular Object Pose Estimation}
Many recent works explore object pose estimation beyond the standard scenarios of datasets that show hand-sized objects at limited distance, and with limited variability regarding illumination~\cite{hu2021wide,viviers2024advancing,sapienza2023underwater,tang2024rov6d,sun2023panelpose,ulmer2024orbital,monguzzi2024cable}.
The pose estimation of satellites in orbit is gaining momentum, which is in conjunction with European Space Agency (ESA) endeavours for robotic space debris removal~\footnote{https://www.esa.int/Space\_Safety/ESA\_purchases\_world-first\_debris\_removal\_mission\_from\_start-up}.
Due to the challenging scenarios in space, RGB is the preferred modality, and challenges are large depth and extreme illumination ranges~\cite{park2023satellite,hu2021wide,kisantal2020satellite,jawaid2023towards,wang2022spacenet,bechini2024robust,wang2023bridging,ulmer2024orbital}.
Concurrently to space, learning-based pose estimation research also starts exploring underwater for robotic applications~\cite{joshi2020deepurl,nielsen2019evaluation,sapienza2023underwater,tang2024rov6d}.
Since depth estimation requires specific working principles or modified systems of depth sensors~\cite{fan2023development}, RGB is advantageous for unconstrained underwater pose estimation from RGB~\cite{tang2024rov6d,sapienza2023underwater}.
Additionally, specific object types and the estimation of their configuration with respect to the environment calls for pose estimation solutions developed for the robotic application~\cite{sun2023panelpose,monguzzi2024cable}.
In summary, while dataset performance on conventional datasets (Section~\ref{sec:datasets}) seems to saturate, robotic application expands to specific scenarios that are beyond the closed-world assumption.
As a consequence, robotics demands specific solutions which will benefit from the information-rich, and well researched RGB modality and foundation models~\cite{firoozi2023foundation}. 

\section{Future Challenges}
\label{sec:future}

The previous section presented popular ongoing research topics.
In the following, we identify gaps in the existing problem landscape based on this overview.
By accumulating these missing problems, we are able to derive high-level challenges that need to be addressed to effectively advance the robotics.

\subsection{Object Ontology}

\begin{figure}[t!]
   \centering
    \includegraphics[width=1.0\columnwidth]{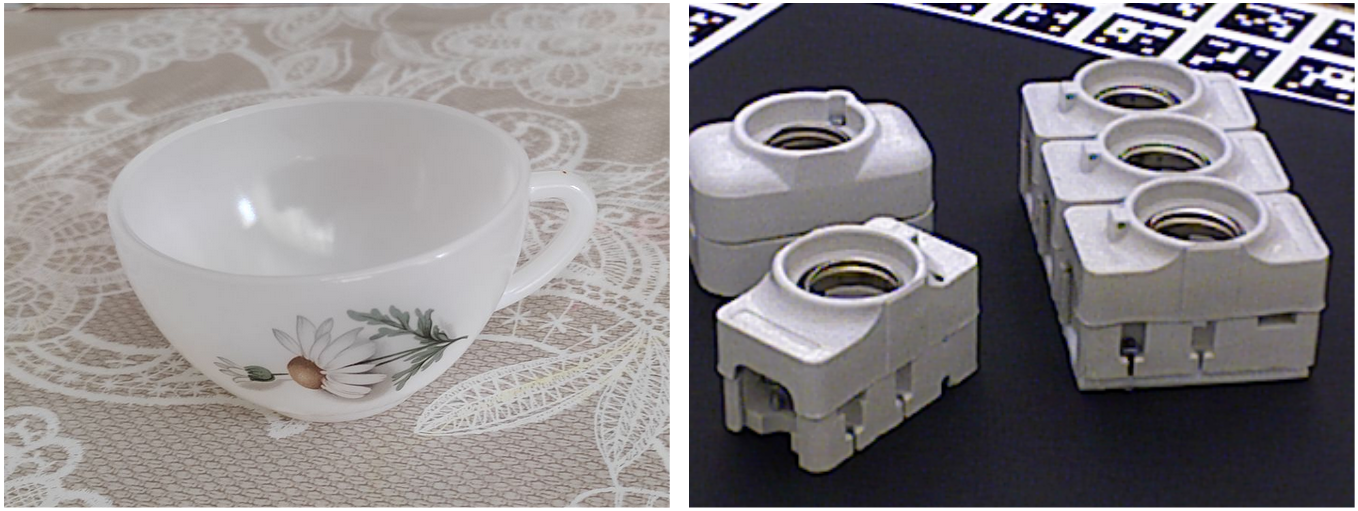}
   \caption{\textbf{Object ambiguities.} The left image shows a cup that exhibits similarities to the categories \textit{cup} and \textit{bowl} of DREDS, Figure~\ref{fig:category_viz}. The right image shows an example where one object is a part of another.}
   \label{fig:ontology}
      \vspace{-2ex}
\end{figure}

A topic largely untouched and completely unsolved is how to estimate poses of objects which are excluded from the known categories and for which no prior information like CAD-models or training images are available.
Novel and category-level pose estimation assumes prior information about the object origin, which is used for prediction making.
Again, coming back to the open-world scenario, generalizing to out-of-distribution objects poses a number of problems.
How to incorporate the pose of truly unknown objects into vision pipelines, if no prior information is available?
Category-level approaches interpolate between training samples of known categories. What if unknown objects lie at the intersection of two categories. 
Consider the left image of Figure~\ref{fig:ontology}, The shape of the cup is similar to the instances of DREDS's category \textit{bowl}, yet has a handle like the instances of the category \textit{cup}, see Figure~\ref{fig:category_viz}.
While \textit{bowl} has to be treated as rotationally symmetric, since missing cues for disambiguation, \textit{cup} is not symmetric. 
How to handle such scenarios and prevent interpolating between the two, or possibly more, object origins of categories that are matched?
Another related issue are objects that are parts of others.
An example is visualized in the right image of Figure~\ref{fig:ontology}. 
TLESS object $6$ is a part of object $7$. 
A case which is difficult to resolve if the ontology of objects is unknown.

Large language models (LLM) and vision-language models (VLM) encode latent object ontologies, thereby enabling generalization to novel instances of known, or even unknown categories~\cite{wu2023tidybot,padalkar2023open,corsetti2023open}.
Verification of LLM reasoning using formal ontologies may enhance the robustness and trustworthiness of learned ontologies, ensuring adherence to industry and robotic standards~\cite{tenorth2009knowrob}. 

\subsection{Deformable and Articulated Objects}
\label{sec:obj_deform}

Most common methods and datasets for monocular pose estimation rely on the assumption of rigidity, where the shape remains fixed and only the position and orientation changes.
Pose estimation of deformable and articulated objects, however, presents significant challenges due to their variability in shape.
Articulated objects may be treated as individual object parts that are connected by joints, hence only introducing a limited number of additional degrees of freedom~\cite{EisnerZhang2022FLOW} as compared to rigid objects.
The method in~\cite{liu2022toward} assumes a finite set of joints and uses NOCS~\cite{wang2019normalized} for local geometric correspondence prediction on each part.
While this presents an intuitive solution for non-rigid objects with only a few degrees of freedom, transferring such a strategies to deformable objects seems infeasible due to their shape's infinite degrees of freedom. The partitioning of the object into sub-entities typically degenerates into dense prediction of its surface or volume.
As a consequence, approaches for handling deformable objects are commonly task specific and may involve robotic manipulation to incrementally learn the object morphology~\cite{corl2020softgym,chi2021garmentnets}.
Yet, handling these types of objects is of particular interest, since they are common in everyday life, e.g. bags and textiles, and thus are important for robotics. 

Objects that undergo smaller scale deformations may be handled by estimating local correspondences~\cite{florence2018dense,goodwin2022zero}.
Such approaches allow to treat such objects within the existing pose estimation framework by estimating dense geometric correspondences from the local neighborhood. 
Alternatively, with robotics-based approaches for textile handling in mind, reasoning about an object's deformability might require solving tasks like learning semantically meaningful parts~\cite{chen2023autobag}.
Apart from learning dense displacement from a rest shape, there is no clear common definition of a deformable object's ``pose'' and origin. Therefore different metrics are used for evaluation in related works, complicating comparison.
Evaluating on proxy tasks like grasping is valid to show the efficacy for that task, yet metrics and formal definitions are required for general quantitative evaluation, reproducibility, and comparability.

Handling arbitrary objects requires algorithms to understand objects from a global, topological and a local, geometrical point of view.
Factors describing the change of physical configuration under deformation need to be encoded in order to effectively extrapolate to out-of-distribution objects.

\subsection{Scene-level Consistency}

Incorporating geometrical feedback loops for pose estimation leads to performance improvements through joint pose refinement of multiple objects~\cite{aldoma2012global,sock2018multi,labbe2020cosypose} and verification of physical plausibility~\cite{bauer2020verefine}. By considering multiple objects in the scene in parallel, their mutual occlusion and support relationships allow to restrict the space of admissible poses and, thereby, simplifies the pose estimation task.

Since the objects' distance from the camera is directly observable in depth images and hence more accurately predicted~\cite{wang2019densefusion}, recent approaches explore depth and occupancy estimation from monocular input~\cite{yen2021inerf,huang2022neural,cao2022dgecn}. While depth sensors produce noisy or incomplete observations of transparent and reflective objects, RGB-based methods are able to reconstruct complete depth observations~\cite{chen2022clearpose}. Depth data is found to generalize well and require less data than RGB methods~\cite{bauer2022sporeagent}. However, this is traded for inability to discern geometrically ambiguous but texturally distinct symmetries. Therefore, starting from RGB input (with optional depth estimation) is expected to cover more challenging situations than using depth sensors alone.
Alternative to depth estimation, end-to-end training detailed in Section~\ref{sec:e2e}, enables similar geometrical guidance of the learning process. 
Regressing multiple pose representations simultaneously enables enforcing consistency between them~\cite{thalhammer2022cope,di2021so}.

A promising direction for future work is to consider the consistency between an estimated scene configuration and, for example, its simulation, reconstruction or rendering. The latter may furthermore incorporate factors such as surface materials which would be especially relevant for reflective and refractive objects in clutter, as their appearance depends on the broader scene. We conjecture that inverse rendering or NeRF-like \cite{mildenhall2021nerf} approaches will advance pose estimation in such situations.

\subsection{Benchmark Realism}

\begin{figure}[t!]
   \centering
    \includegraphics[width=0.8\columnwidth]{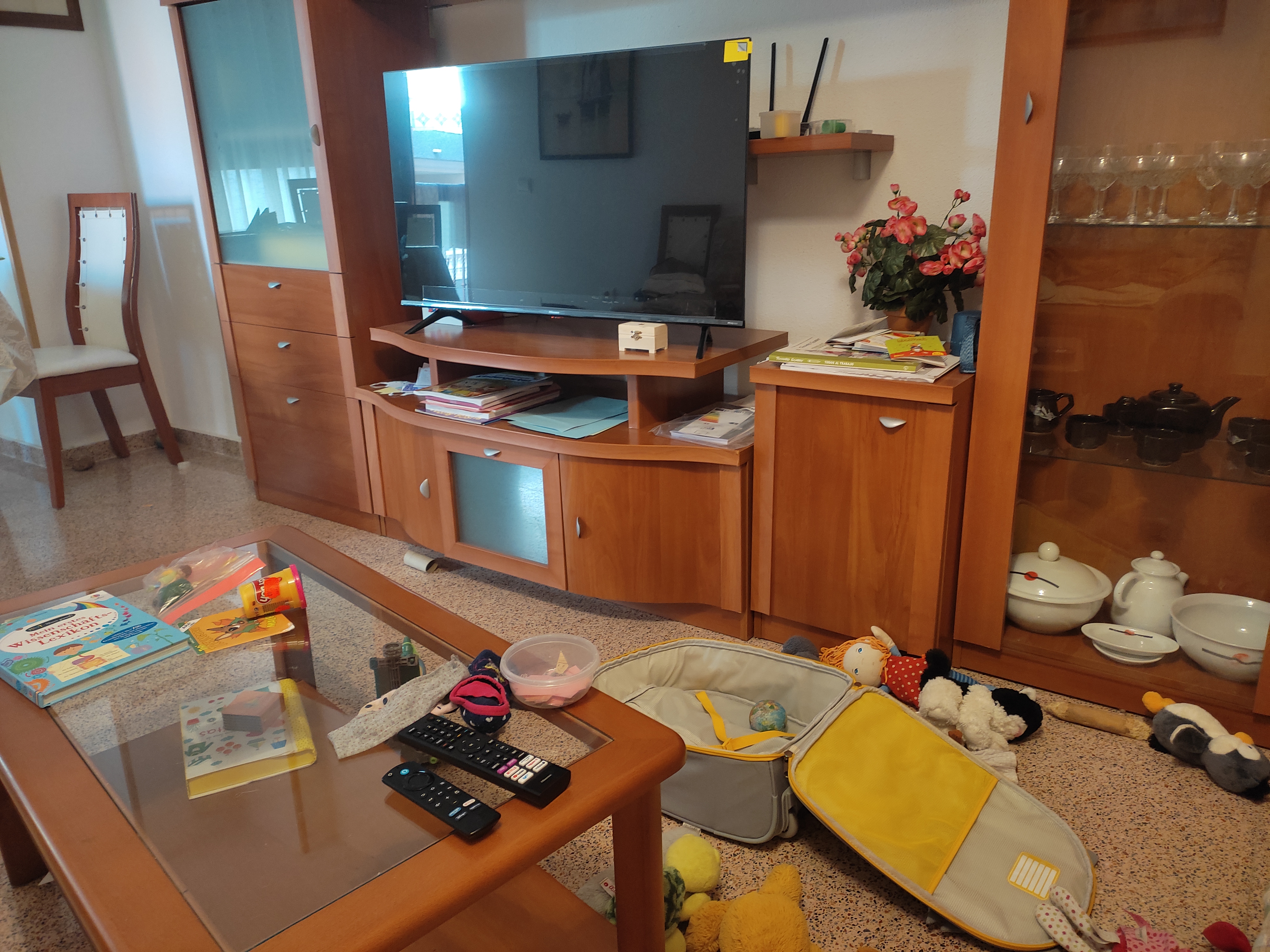}
   \caption{\textbf{Real household scene} featuring diverse challenging conditions including rigid and deformable objects, transparent and opaque objects, some of them behind glass, a large variation in object sizes and placement with respect to the camera, with the view not centered on a single support plane.}
   \label{fig:real_data}
      \vspace{-2ex}
\end{figure}

Existing datasets are still in an early stage with respect to their complexity and challenges in comparison to the real world.
There are a number of aspects which are simplified for the ease of dataset creation and annotation.
These aspects are scene setup, object variation, interactions between objects and varying material properties. 
The following are examples where existing datasets do not depict real-world complexities:
\begin{itemize}
 \item \textbf{Clutter:} Rich multi-object, multi-instance scenarios are required to exhaustively evaluate and improve occlusion handling, and as such general performance.
 Existing datasets, however, only feature manually crafted scenes. As a result, objects of interest are generally placed in the middle of the scene (and the camera view) and the diversity of occluding distractors, as well as their placement, is limited.
 This presents a large discrepancy to real scenarios. Considering for instance a kitchen scenario, occluding objects are unrestricted, and can be as indiscriminate as vegetable scraps. Multiple instances of the same object are apparent in varying states and often interact with one another, their poses are only restricted by the scene geometry, and objects may even appear behind refractive occluders such as refrigerator shelves.
 \item \textbf{Backgrounds:} Relevant datasets mostly feature objects arranged on a single support plane, with a limited set of background textures and non-dataset distractors (see Figure~\ref{fig:instdata}). 
 Only few datasets assume more complex scene backgrounds~\cite{tyree20226,rennie2016rutgers}.
 Yet, generally, datasets that incorporate multi-faceted object placements with respect to lateral and vertical placement in the world and significantly changing scene backgrounds are missing.
 \item \textbf{Object variations:} As stated in Section~\ref{sec:obj_material}~\ref{sec:obj_deform}, mainstream approaches' ability to handle object property variation is very limited. 
 This is a consequence of the standard datasets that typically feature a single type of material (e.g., textured, metallic, refractive), and similar-sized and rigid objects.
 
 In contrast, a realistic dataset needs to feature a diverse object ontology. Objects need to be a) annotated on instance- and category level, and novel objects and categories, which are not available for training, need to be introduced for testing. b) Object sizes need to vary from small to large, and shapes from simple to complex. c) Visual materials need to include opaque and transparent, diffuse and reflective, and textured and texture-less variations. These dimensions are ideally quantified with tractable metrics to allow creating exhaustive test scenarios and to quantify pose estimation accuracy.
\end{itemize}

Figure~\ref{fig:real_data} shows a real household scene, with different objects of various sizes and materials, some also behind glass and as multiple instances. 
Objects are distributed over the full image space, many objects are further away from the camera as is assumed by most of the standard datasets, and the image is not centered with respect to a support plane.
In order to improve dataset creation, a paradigm shift needs to happen.
For one, the dataset creation and annotation process itself needs to improve.
The standard process is to choose an object set and subsequently deliberately place them in backgrounds that allows pose annotation, respectively simplifies annotation, e.g. marker boards as done in~\cite{chen2022MP6D,hinterstoisser2012model,doumanoglou2016recovering,homebrewedDB,hodan2017tless,wang2019normalized,liu2020keypose,fang2022transcg}.
Rather, test scenarios for household robotics should organically grow and objects of interest are ideally chosen from the set of objects that happened to be there.
This way, not only the object arrangements, but also the occlusion patterns are more natural, and ultimately, datasets created in that fashion depict the real-world complexity and object variations more accurately.
To this end, recent annotation tools allow to propagate poses over consecutive frames, given camera poses are accurate enough~\cite{suchi20233d}. 
The lower bound for pose annotation error is thus limited by camera pose accuracy. Still, object instances that are strongly occluded will be difficult to annotate.
Geometric reasoning for objects and the scene during annotation is required to alleviate such issues.

\subsection{Environmental Impact}

Object pose estimation research and application consumes large amounts of energy.
Especially under current concerns regarding sustainable research and environmental impact, addressing this issue is an obligation for future research.

Using instance-level pose estimators is still the standard case for robotic scenarios. 
This requires re-training them every time a new object or object set is used.
For example, the pose estimator of GDR-Net~\cite{wang2021gdr} trains on average approximately $6h$ on an NVIDIA RTX 3090 (with a TDP of 350W) for maximizing its performance for a single object of the BOP-challenge.
This requires more than $8$ days of continuous GPU usage for the $33$ objects of HB, or about $69$kWh for the GPU alone (assuming full utilization) -- the amount of energy the average U.S. consumer used over $2.1$ days, or the worldwide-average consumer used over $7.9$ days in 2021\footnote{https://www.eia.gov/international/data/world/electricity/electricity-consumption. Most recent numbers (2021) accessed on July 20, 2023.}.
Category-level object pose estimation may have a lower environmental impact, since trained models generalize to objects of the same category and as such may be used in more application scenarios, but still models are usually trained for each category separately.
Both these types of approaches exhibit fast inference, in the range of seconds, but are very limited with respect to the objects for which poses are inferred.
Novel object pose estimators have the potential to alleviate the need for long training times~\cite{goodwin2022zero,fan2023pope}, albeit, most approaches are still in the range of hours to days~\cite{thalhammer2023self,shugurov2022osop,labbe2022megapose}. Runtime is usually in the range of minutes since template matching requires calculating similarities to thousands of templates.
As such, novel object pose estimation still has not reached a state that largely improves power consumption.

Exploring alternatives to template matching might reduce the runtime for novel object pose estimation.
For instance, methods for inference from a few or single reference images are a potential solution, although currently still lack accuracy.
An alternative is offered by foundational models such as ViTs, pre-trained on large datasets, which already showed that they transfers well to pose estimation~\cite{goodwin2022zero,fan2023pope,thalhammer2023self}.
Exhaustively exploring such approaches that require no re-training for novel objects or changing test domains has the potential to reduce the power consumption of research and for robotic application, as the power expanded for the pre-training stage, while larger, is shared across all final applications.
Generally, methods that require no training and exhibit fast inference are important to allow easy and resource efficient utilization for diverse scenarios.

\subsection{Generalist Object Manipulation}

Many robotic tasks in, e.g., industrial or household environments, may be seen as different instances of rearrangement~\cite{batra2020rearrangement}, often solvable by pick-and-place actions and thus connected to grasping. While grasping is often approached via object geometry, i.e., by grasp or object pose estimation, diverse alternative research directions exist, such as universal grasping~\cite{mahler2017dex,sundermeyer2021contact,fang2020graspnet,fang2023anygrasp}, generalist robot policy learning~\cite{padalkar2023open,team2023octo}, and foundation models~\cite{firoozi2023foundation,black2023zero}.
Universal object grasping is an effective alternative to grasping based on poses, and even has benefits.
Most approaches that reason about the graspability and the grasp's location do generalize tremendously well over object types~\cite{mahler2017dex,sundermeyer2021contact,fang2020graspnet,fang2023anygrasp}.
Such methods show grasp generalization to completely novel geometries without requiring priors at inference, which is a major shortcoming of object pose estimation.
However, object grasping based on poses enables complex pick and place applications since transformations to the target location can be inferred.
One family of approaches that is however solving such cases are generalist robot policies~\cite{padalkar2023open,team2023octo}.
They leverage the joint embedding of VLM to perform object manipulation without explicit pose knowledge.
Recent works showcase that complex manipulation tasks can be performed, such as object stacking and arranging objects relative to others. 
Especially interesting to robotics is the capability of handling multiple input modalities, for example combining vision with language instructions or sound, and controlling different robot embodiments and kinematic structures, such as robotic arms and grippers with different numbers of joints.
These advancements are possible through the rise of foundation models like~\cite{jiang2023mistral}.
The equivalent for vision~\cite{oquab2023dinov2,zhang2024tale} is shown to allow zero-shot execution of vision and robotic tasks~\cite{firoozi2023foundation,black2023zero,kirillov2023segment}.
The combination of visual image description~\cite{radford2021learning} and NeRF or Gaussian Splatting enables encoding language into reconstructions and thus allows object pose querying with certain generalization capabilities~\cite{kerr2023lerf}.

Object pose estimation for robotics is still useful for scene understanding and complex manipulation tasks that require high accuracy.
However, trends are recognizable that indicate that object localization, and thus manipulation, is to some extent already happening in encoding space~\cite{padalkar2023open,team2023octo}
This is a very promising and large step toward human-like perception which does not require explicit object poses to enable the manipulation of objects.

\section{Conclusion}
\label{sec:conclusion}

This work identifies and presents current research problems for monocular single-shot 6D object pose estimation and derives important future challenges. 
A discrepancy between object pose estimation research, that aims at maximizing performance on benchmark datasets, and robotic requirements is uncovered.
Consequently, we present important future research challenges toward bringing both fields closer together.
Benchmarking datasets need to exhibit more realism, in order to speed up computer vision research for robotics.
Having datasets with diverse known and unknown objects in known and unknown categories allows searching for ontologies describing objects on a fundamental level.  
Rich multi-object scenarios with different challenging surface materials and deformations requires advancing the state of the art for handling these properties and enable effective implementation of robots in households.
Raising pose estimates to configurations that are consistent on scene-level will provides reliable priors for robotic downstream tasks. 
Especially deformable and articulated objects require novel metrics to measure benchmarking progress.
And, finally, climate change and sustainable research demands reducing the ecological footprint of algorithms.

\section*{Acknowledgments}
We gratefully acknowledge the support of the EU-program EC Horizon 2020 for Research and Innovation under grant agreement No. 101017089, project TraceBot, the Austrian Science Fund (FWF), under project No. I 6114, project iChores, under project No. J 4683, project Making Sense of Objects, and the NVIDIA Corporation.

This work has been accepted for publication in IEEE Transactions on Robotics. The final version can be accessed at: \url{https://ieeexplore.ieee.org/document/10609560}, Digital Object Identifier: 10.1109/TRO.2024.3433870.

{\small
\bibliographystyle{IEEEtran}
\bibliography{root}
}

\vfill

\end{document}